\documentclass{article}

\PassOptionsToPackage{numbers, compress}{natbib}

\usepackage[preprint]{neurips_2024}

\usepackage{amsmath}
\usepackage{amssymb}
\usepackage{mathtools}
\usepackage{amsthm}

\usepackage{graphicx}
\usepackage{amsmath}
\usepackage{amssymb}
\usepackage{booktabs}
\usepackage{adjustbox}
\usepackage{boldline}
\usepackage{pifont}
\usepackage{xcolor}

\usepackage[utf8]{inputenc} %
\usepackage[T1]{fontenc}    %
\usepackage{hyperref}       %
\usepackage{url}            %
\usepackage{booktabs}       %
\usepackage{amsfonts}       %
\usepackage{nicefrac}       %
\usepackage{microtype}      %
\usepackage{xcolor}         %
\usepackage[capitalize,noabbrev]{cleveref}
\crefname{appendix}{Appx.}{Appx.}
\crefname{equation}{Eq.}{Eqs.}
\crefname{figure}{Fig.}{Figs.}
\crefname{tabular}{Tab.}{Tabs.}
\crefname{section}{Sec.}{Secs.}
\crefname{table}{Tab.}{Tabs.}

\theoremstyle{plain}

\theoremstyle{definition}

\theoremstyle{remark}

\usepackage{wrapfig,lipsum,booktabs}

\usepackage{multirow}
\usepackage{slashbox}
\usepackage{diagbox}
\usepackage{dsfont}
\usepackage{algorithm}
\usepackage{algpseudocode}
\usepackage{algorithmicx}
\usepackage{enumitem}
\usepackage{caption}
\usepackage{subcaption}
\usepackage{abdalmageed_style}
\algrenewcommand\algorithmicrequire{\textbf{Input:}}
\algrenewcommand\algorithmicensure{\textbf{Output:}}

\usepackage{newpxtext}
\makeatletter
\newcommand{\ie}{\emph{i.e.}\@ifnextchar.{\!\@gobble}{}}
\newcommand{\eg}{\emph{e.g.}\@ifnextchar.{\!\@gobble}{}}
\newcommand{\etc}{etc\@ifnextchar.{}{.\@}}
\makeatother

\title{DiffusionCounterfactuals: Inferring High-dimensional Counterfactuals with Guidance of Causal Representations}

\author{
Jiageng Zhu$^{1,3}$ \qquad Hanchen Xie$^{2,3}$ \qquad Jiazhi Li$^{1,3}$ \AND
   \qquad Wael AbdAlmageed$^{1,2,3}$  \\
$^1$ USC Ming Hsieh Department of Electrical and Computer Engineering \\
$^2$ USC Thomas Lord Department of Computer Science\\
$^3$ USC Information Sciences Institute \\
$^4$ Holcombe Department of Electrical and Computer Engineering, Clemson University \\
\tt\small  \{jiagengz, hanchenx, jiazhil\}@isi.edu \qquad wabdalm@clemson.edu}

\begin{document}

\maketitle

\begin{abstract}

Accurate estimation of counterfactual outcomes in high-dimensional data is crucial for decision-making and understanding causal relationships and intervention outcomes in various domains, including healthcare, economics, and social sciences. However, existing methods often struggle to generate accurate and consistent counterfactuals, particularly when the causal relationships are complex. We propose a novel framework that incorporates causal mechanisms and diffusion models to generate high-quality counterfactual samples guided by causal representation. Our approach introduces a novel, theoretically grounded training and sampling process that enables the model to consistently generate accurate counterfactual high-dimensional data under multiple intervention steps.  Experimental results on various synthetic and real benchmarks demonstrate the proposed approach outperforms state-of-the-art methods in  generating accurate and high-quality counterfactuals, using different evaluation metrics.

\end{abstract}

\section{Introduction}
\label{sec:introduction}

Causality, the study of cause-and-effect relationships, enhances our understanding of complex systems, as well as our ability to manipulate them. Learning causality form data enables the prediction of outcomes and the understanding of underlying mechanisms \cite{reason:Pearl09a}. This is crucial when decisions and interventions require more than just correlations \cite{bib:towards-causal}. 
Causal representation learning (CRL) methods \cite{DBLP:conf/cvpr/YangLCSHW21,bib:do-VAE} have made significant progress in this direction, enabling models to represent causal relations from high-dimensional data.

Counterfactual inference involves estimating hypothetical scenarios for a given factual observation under specific interventions \cite{reason:Pearl09a}. An intervention is an action taken to alter causal factors to observe their effects on other factors in a causal mechanism.
Counterfactual inference is crucial in domains such as healthcare \cite{counterfactual-medicine}, economics \cite{counterfactual-economics}, and policy-making \cite{counterfactual-policy}, where understanding the impact of interventions is essential for informed decision-making. However, counterfactual inference becomes more challenging when dealing with high-dimensional data, such as images or complex time series, due to the complex relationships among numerous variables. Therefore, there is a growing need for methods that can effectively perform counterfactual inference in high-dimensional spaces, enabling the exploration of hypothetical scenarios in complex, real-world situations.

Existing causal representation learning methods, such as CausalVAE \cite{DBLP:conf/cvpr/YangLCSHW21} and CausalGAN \cite{kocaoglu2018causalgan}, often struggle to achieve satisfactory results when estimating counterfactual outcomes in high-dimensional data, such as images or complex time series \cite{bib:do-VAE}.
The limitations of these methods in accurately modeling the complex causal relationships underscore the need for more sophisticated approaches. Meanwhile, diffusion-based generative models \cite{ddpm,ddim,guided-diffusion,ldm}  have recently been popular for improving the quality of generates images with unparalleled ability to craft detailed and diverse data simulations. Diffusion models are trained to gradually refine noise into structured outputs, demonstrating an impressive ability to generate complex data patterns, which makes them a potentially  promising alternative to address the limitations of counterfactual reasoning methods.

\begin{figure*}
    \centering
     \includegraphics[width=0.92\textwidth]{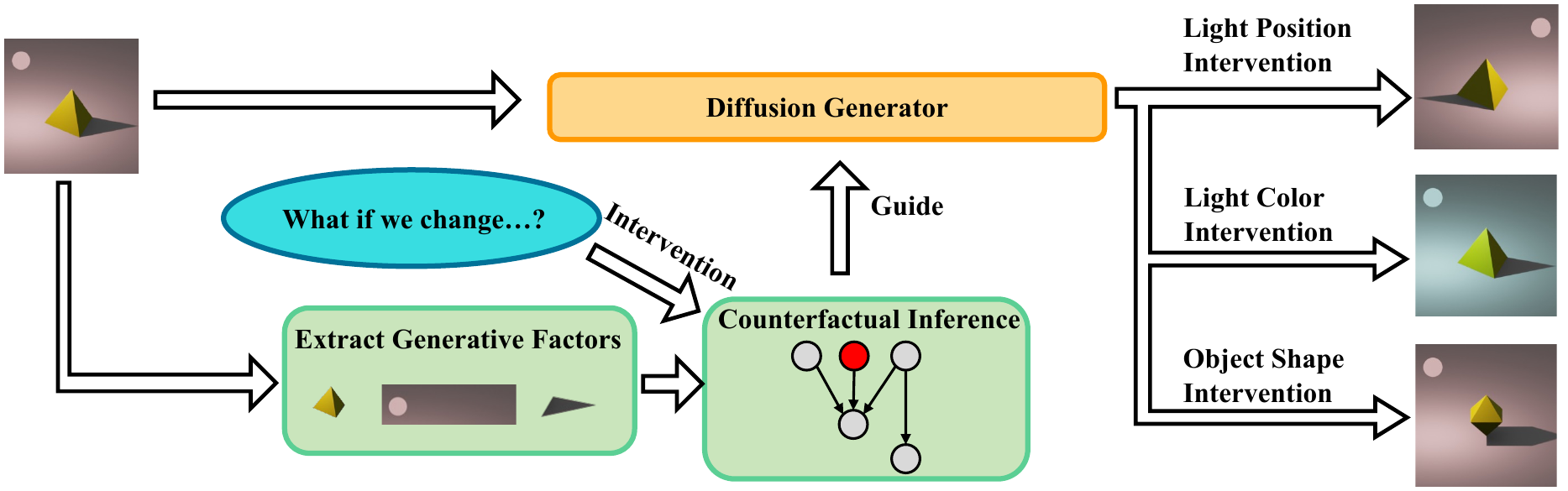}
    \caption{Guided by the causality, the diffusion generation can answer “what if” questions, i.e., counterfactual inference, in the input space.} %
    \label{fig:teaser}
    \vspace{-1\baselineskip}
\end{figure*}
Therefore, to address the limitations of existing counterfactual inference methods, we propose utilizing the generative ability of diffusion models guided by causal representations. We conjecture that controlling the diffusion process with the guidance of causal representations and the underlying causal mechanisms will significantly improve counterfactual inference.
Integrating causal inference mechanisms into diffusion-based generative models represents a fundamental shift, enabling these models to understand and manipulate the causal factors underlying the data. This shift is crucial for tasks where interpreting counterfactuals is essential, such as medical diagnosis \cite{medical_cf}, and social strategic planning \cite{social_cf}.

To achieve this goal, the proposed approach seeks to leverage the strengths of diffusion models in handling high-dimensional data while empowering them with the ability to understand and model the causal structures underlying the data. 
Specifically, Our proposed DiffusionCounterfactuals framework presents a promising advancement by guiding the diffusion generation process using causal representations and their associated causal mechanisms. DiffusionCounterfactuals introduces an innovative training process that concurrently learns to reconstruct high-quality images and uncover the underlying causal mechanisms controlling the generative factors. During training, the diffusion model generates noisy versions of the input images, while a causal projector predicts the generative factors from these noisy inputs. Additionally, a Neural Structural Causal Model (NSCM) is employed to learn the causal relationships among the generative factors, ensuring the model captures the true causal structure. For counterfactual inference, DiffusionCounterfactuals introduces a modified sampling process that incorporates a gradient-based guidance term derived from the NSCM and causal projector. This guidance term encourages the generated counterfactual images to align with the desired interventions on the causal factors, producing a more accurate and causally consistent counterfactuals. Furthermore, DiffusionCounterfactuals employs a novel self-adjusted scalar for guidance, which dynamically adapts the strength of the guidance term based on the timestep and the uncertainty in the diffusion process, improving the quality and consistency of the generated counterfactuals.

The main contributions are (1) DiffusionCounterfactuals: a novel framework integrating diffusion models and causal representation learning to generate accurate, causally consistent counterfactuals in input space.
(2) A novel training process that simultaneously reconstructs high-quality images and discovers underlying causal mechanisms governing generative factors.
(3) A modified sampling process incorporating gradient-based guidance and a self-adjusting scalar to generate counterfactuals consistent with desired interventions on causal factors.
(4) Extensive empirical evaluations on various datasets, demonstrating that DiffusionCounterfactuals outperforms state-of-the-art methods in generating high-quality and causally consistent counterfactuals.

\section{Related work}

\vspace{-0.3\baselineskip}
\noindent \textbf{Diffusion Generative Models}
Diffusion generative models have recently gained attention for generating high-quality samples in various domains, including images \cite{ddpm}, audio \cite{diffusion-audio}, and text \cite{diffusion-text}. These models learn a reverse diffusion process that gradually denoises a sample from random noise \cite{first-diffusion}. Sohl-Dickstein et al. \cite{first-diffusion} introduced diffusion probabilistic models, which learn a Markov chain of diffusion steps to transform complex data into simple noise distributions, demonstrating their effectiveness in generating high-quality images. Ho et al. \cite{ddpm} proposed Denoising Diffusion Probabilistic Models (DDPMs), improving sample quality and diversity by estimating data distribution gradients at each step. Song et al. \cite{ddim} introduced Denoising Diffusion Implicit Models (DDIMs), providing an efficient and flexible framework for controlling the quality-diversity trade-off. Dhariwal and Nichol \cite{guided-diffusion} showcased the superior performance of diffusion models compared to generative adversarial networks (GANs) \cite{gan}. Despite their success, applying diffusion models to counterfactual estimation remains challenging, as incorporating causal knowledge and ensuring consistency with desired interventions is an open problem \cite{diffusion-explain}. Sanchez et al. \cite{diffscm} propose classifier guided diffusion \cite{guided-diffusion}  can be viewed as categorical label (e.g., class) guided counterfactual inference. 

\noindent \textbf{Causal Representation Learning}
Causal representation learning aims to discover the underlying causal structures and relationships from observational data \cite{bib:towards-causal}. By learning causal representations, models can capture the intrinsic causal mechanisms governing the data generation process and enable counterfactual reasoning \cite{kocaoglu2018causalgan,DBLP:conf/cvpr/YangLCSHW21}. CausalVAE \cite{DBLP:conf/cvpr/YangLCSHW21} is a prominent approach that combines variational autoencoder (VAE) \cite{DBLP:conf/iclr/HigginsMPBGBML17}  with causal inference principles to learn causal representations. CausalVAE employs a structured latent space that reflects the causal relationships among variables, allowing for the generation of counterfactuals by intervening on specific latent factors. CausalGAN \cite{kocaoglu2018causalgan} is another notable approach that integrates causal modeling with generative adversarial networks (GANs) \cite{gan}. CausalGAN learns a causal implicit generative model by utilizing a structured generator network that captures the causal relationships among variables. The generator is trained adversarially against a discriminator that distinguishes between real and generated samples, ensuring the generation of realistic counterfactuals. While CausalVAE and CausalGAN have shown promising results in learning causal representations, they may often struggle with generating high quality high-dimensional and complex data \cite{avoid-discrimination}.

\section{DiffusionCounterfactuals}
\label{sec:causal-diffusion}

\vspace{-0.3\baselineskip}
\subsection{Problem Formulation}
\vspace{-0.2\baselineskip}
\label{sec:setup}

We begin by defining the causal generation process and give 
defintion of the counterfactual inference in the input space. Given a set of observed inputs 
$\mathcal{X} = \{x^{(1)}, x^{(2)}, ..., x^{(n)}\}$, we assume each image $x^{(i)} \in \mathbb{R}^{d_x} $ is generated by a set of generative factors $\mathcal{Z}^{(i)} = \{z^{(i)}_1, z^{(i)}_2, ..., z^{(i)}_{d_z}\}$ through an ideal generation model $g^*$, i.e., $x^{(i)} = g^*(\mathcal{Z}^{(i)})$. The causal relationships between generative factors are governed by a structural causal model (SCM) $f$, which is defined as $f_i: \xi_i \times \prod_{j \in pa_i} z_j \rightarrow z_i, \forall i \in {1, 2, ..., d_z}$ \cite{reason:Pearl09a}, where function $f_i$ describes how a factor $z_i$ is determined by its causes, i.e., parents ($pa_i$), and $\xi_i$ is an independent exogenous random variable.  The entire SCM functions entail a faithful causal structure $\mathcal{G}$  which can be represented using directed acyclic graph (DAG) \cite{reason:Pearl09a}, in which all independencies \cite{reason:Pearl09a} (\ie, if $A \perp B | C$ in $\mathcal{G}$, then $P(A|B,C) = P(A|C)$) are encoded.

\begin{figure*}
    \centering
     \includegraphics[width=\textwidth]{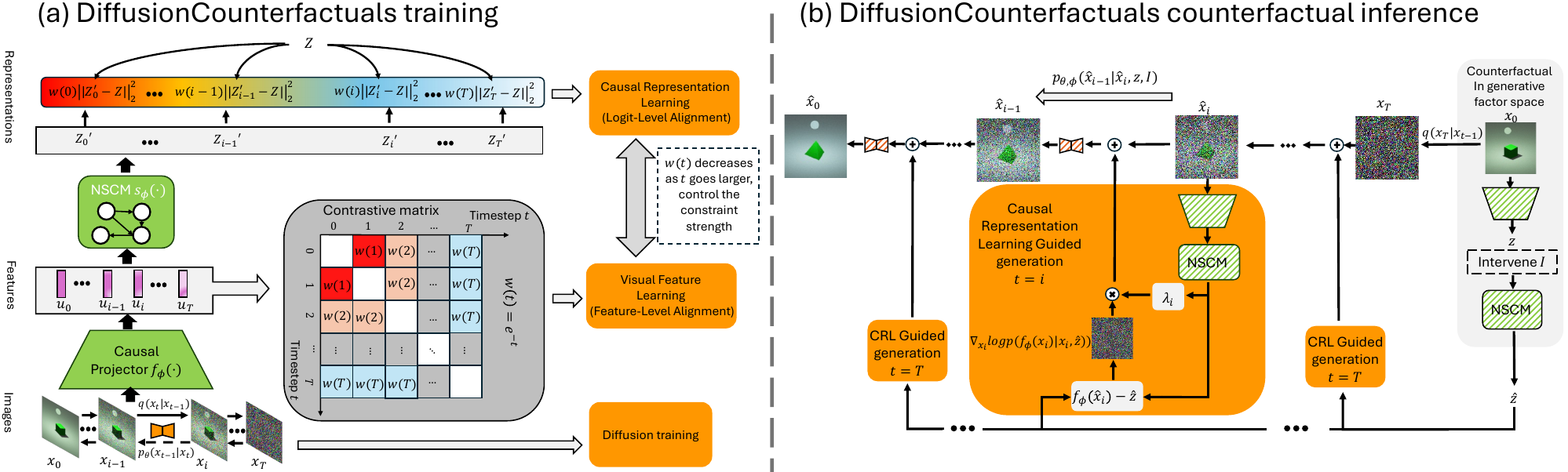}
    \caption{DiffusionCounterfactuals Framework: Training and Counterfactual Inference.
(a) The DiffusionCounterfactuals training process learns a diffusion reconstruction and discovers the causal factors along with their corresponding causal relations.
(b) The counterfactual inference process generates counterfactual images by conditioning on estimated intervention outcomes of specific factors in the generative factor space, using the learned model. } %
    \label{fig:framework}
    \vspace{-\baselineskip}
\end{figure*}

Counterfactuals in the input space are defined such that, given a factual input $x^{(j)}$, the counterfactual $\hat{x}^{(j)}$ is generated through $\hat{x}^{(j)} = g^*(\hat f(\mathcal{Z}^{(j)}))$, where $\hat f$ represents the new causal mechanism with intervened factors. Meanwhile, counterfactuals in the generative factors space are defined such that for an instance $\mathcal{Z}^{(i)}$, by applying intervention $\mathcal{I}$ on some factors, the model generates hypothetical scenarios $\hat{\mathcal{Z}}^{(i)}$ \cite{reason:Pearl09a}.

\subsection{Architecture}
\label{sec:framework}
To enable causal-aware controlled generation in diffusion models, we propose DiffusionCounterfactuals, which is illustrated in \cref{fig:framework}. During learning, the model simultaneously focuses on image reconstruction and causal representation learning (CRL). The diffusion model learns to maximize $\log p(x)$, while the causal projector and Neural Structural Causal Model (NSCM) learn to maximize $\log p(\mathcal{Z}|x)$. This maximizes the joint distribution $\log p(x, \mathcal{Z})$, according to Bayesian theory: $\log p(x, \mathcal{Z}) = \log p(\mathcal{Z}|x) + \log p(x)$.

During the counterfactual inference, in addition to the standard diffusion reverse sampling, we incorporate the gradients backpropagated from the NSCM $s_{\phi}$ and a causal projector. This gradient is derived from the loss function that takes the outputs of $s_{\phi}$ as inputs. We discuss the training details in \cref{sec:training}, causal-aware counterfactual inference in the input space in \cref{sec:sampling}, and the self-adjusting scalar for enhancing generation quality in \cref{sec:adjusted-scalar}.

\subsection{Training}
\label{sec:training}
As described in \cref{sec:framework}, we utilize a neural network (NN) to learn the joint distribution $p(x, \mathcal{Z})$ and discover the underlying causal mechanism simultaneously. The entire model includes a generator $g_\theta$, causal projector $h_{\phi}$ and a Neural  Structural Causal Model (NSCM) $s_{\phi}$. As illustrated in \cref{fig:framework}, to achieve the training objectives, we incorporate two sub-tasks --- (1) reconstruction task  and (2) causal representation learning (CRL) task.

During training, the model has access to the ground-truth generative factors $\mathcal{Z}^{(i)}$ and the corresponding input samples $x^{(i)}$ \cite{DBLP:conf/cvpr/YangLCSHW21,kocaoglu2018causalgan}. Given a sample $x^{(i)}$, the diffusion model generates a noisy version $x^{(i)}_t$ for a timestep $t$. The generator $g_{\theta}$ is trained to approximate the applied noise using mean square error (MSE) loss. The noisy input $x^{(i)}_t$ is then fed into a causal projector $h{\phi}$ to produce factors that approximate the ground-truth generative factors $\mathcal{Z}^{(i)}$ of the original sample $x^{(i)}$. To discover the underlying causal mechanism and graph structure, the output of the causal projector is further fed into the NSCM $s_{\phi}$ to propagate causal effects from cause factors to effect factors. The causal projector $h_{\phi}$ and NSCM $s_{\phi}$ are trained by minimizing the difference between the ground-truth generative factors $\mathcal{Z}^{(i)}$ and their respective outputs. By minimizing these differences, the causal projector learns to predict generative factors from noisy inputs, and the NSCM discovers the causal mechanism and graph structure while enforcing an acyclic constraint \cite{bib:notears, bib:gae, deci}.

The causal projector is trained using a weighted MSE loss, where the loss from different timestep noisy inputs $x_t$ is fed into $h_{\phi}$. To account for the increasing blurriness of $x_t$ at larger timesteps, a weight function $w(t)$ is utilized to adjust the loss penalty, with smaller penalties applied for larger values of $t$. Additionally, to facilitate the generation process during the reverse diffusion step, which samples from $T$ to $0$ step by step, a contrastive loss (e.g., SimCLR \cite{simclr})
is adopted on the output of the second last layer of $h_{\phi}$. This constrastive loss enables feature-level alignment for noisy inputs at different timesteps that originate from the same original input. Similar to the logit-level loss, a time weight $w_{t_1,t_2}$ is applied to the contrastive loss to control its strength. The training loss of $h_{\phi}$ can be expressed as:
\begin{small}
\begin{equation}
\begin{split}
    L_{h_{\phi}} = &\sum_{i} w(t) ||\mathcal{Z}^{(i)} - h_{\phi}^{(i)}(x_t) ||_2^2 + \sum_{j} w(t)  ||\mathcal{Z}^{(j)} - h_{\phi}^{(j)}(x_t) ||_2^2 \\ &- w(max(t_1^{(i)}, t_2^{(i)})) \log \frac{\exp \left(\operatorname{sim}\left(\boldsymbol{u}{i}, \boldsymbol{u}{j}\right) / \tau\right)}{\sum{k=1}^{2 N} \mathbb{1}{[k \neq i]} \exp \left(\operatorname{sim}\left(\boldsymbol{u}{i}, \boldsymbol{u}_{k}\right) / \tau\right)}
\end{split}
\end{equation}
\end{small}
Here, $w(t) = e^{-t}$ is chosen to control the loss strength at different timesteps.

To discover the underlying causal mechanism as well as the causal structure, we leverage the following constraint to train NSCM:
\begin{small}
   \begin{equation}
\label{eq:NSCM-constraint}
\begin{split}
        & min_{s_{\phi},A} \sum_i ||\mathcal{Z}^{(i)} - s_{\phi}(\mathcal{Z}^{(i)}, A)  ||_F^2 + \alpha ||A||_1 \\ 
    & \text{subject to } H(A) = tr(e^{A \odot A})=0
\end{split}
\end{equation} 
\end{small}
where $A$ is the causal graph to be discovered and $H(A)$ is the DAG constraint \cite{bib:notears,bib:gae,deci}. Utilizing the augmented Lagrangian method, the loss for training the NSCM is defined as:
\begin{small}
    \begin{equation}
\label{eq:NSCM-training}
    L_{s_{\phi}} = \sum_i ||\mathcal{Z}^{(i)} - s_{\phi}(\mathcal{Z}^{(i)}, A)  ||_F^2 + \alpha ||A||_1 + \beta H(A) + \frac{\rho}{2}|H(A)|
\end{equation}
\end{small}
The overall loss, obtained by combining the losses for the diffusion generation, causal projector, and NSCM, is summarized as:
\begin{equation}
\label{eq:total-loss}
    \mathcal{L} = ||\epsilon_t - \epsilon_{\theta}(x_t,t)||_2^2 + L_{h_{\phi}} + L_{s_{\phi}}
\end{equation}

\subsection{Causal Representation Guided Counterfactual Generation}
\label{sec:sampling}
In this section, we derive a modified sampling process for generating counterfactual samples guided by causal representations. The process builds upon standard diffusion model sampling and incorporates an additional guidance term to align generated images with desired causal representations. The modified sampling equation is derived by maximizing the log-likelihood of the desired causal representation given the noisy image at each timestep.  To enforce the reverse diffusion process conditioned on the causal representation $\mathcal{Z}$, we model each step following the approach in \cite{guided-diffusion}, as shown in \cref{eq:generation-likelihood}
\begin{small}
    \begin{equation}
\label{eq:generation-likelihood}
    p_{\theta, \phi}(x_{t-1}|x_t,z) = Kp_{\theta}(x_{t-1}|x_t)p_{\phi}(\mathcal{Z}|x_{t-1})
\end{equation}
\end{small}
where $K$ is a normalizing constant. To approximate $p_{\phi}(z|x_{t-1})$, we can approximate $\log p_{\phi}(z|x_{t-1})$. By applying a first-order Taylor expansion around $x_{t-1}=\zeta$, we obtain:
\begin{small}
   \begin{equation}
    \begin{split}
    \log p_{\phi}(z|x_{t-1}) & \approx \log p_{\phi}(z|x_{t-1})|_{x_{t-1}=\zeta} + (x_{t-1}-\zeta)\nabla_{x_{t-1}}\log p_{\phi}(z|x_{t-1})|_{x_{t-1}=\zeta} \\
    & = C + (x_{t-1}-\zeta)J
\end{split}
\end{equation} 
\end{small}
where $J$ is the Jacobian with respect to $x_{t-1}$. Further, in diffusion model, each reverse sampling $p_{\theta}(x_{t-1}|x_t)$ follows a Gaussian distribution $\mathcal{N}(\zeta, \Sigma)$ \cite{ddpm}.
Thus, when conditioning on the causal representation, the log-likelihood shown in \cref{eq:generation-likelihood} can be approximated as:
\begin{small}
   \begin{equation}
\label{eq:sampling-total}
    \begin{split}
    log(p_{\theta, \phi}(x_{t-1}|x_t,\mathcal{Z})) & \approx -\frac{1}{2}(x_{t-1}-\zeta)^T \Sigma^{-1}(x_{t-1}-\zeta) + (x_{t-1}-\zeta)J + C_1 \\
    &= -\frac{1}{2}(x_{t-1}-\zeta-\Sigma J)^T \Sigma^{-1}(x_{t-1}-\zeta -\Sigma J) + \frac{1}{2} J^T \Sigma J  + C_2 \\
    & = -\frac{1}{2}(x_{t-1}-\zeta-\Sigma J)^T \Sigma^{-1}(x_{t-1}-\zeta -\Sigma J) + C_3 
\end{split}
\end{equation} 
\end{small}
According to \cref{eq:sampling-total}, each conditional generation process can be approximated as a Gaussian distribution with a mean of $\zeta + \Sigma J$ and a variance of $\Sigma$.

To calculate the Jacobian $J$, let $p(f_\phi(x_t)|x_t)$ denote the probability distribution of the causal representation $f_\phi(x_t)$ given the noisy image $x_t$ at timestep $t$. $f_\phi(x_t)$ is concatenate of the causal representation projector and NSCM, and we assume that the residual of the prediction, i.e., $p(f_\phi(x_t)|x_t)-z$, follows a Gaussian distribution with a mean of $0$ and a variance of $\delta^2$. Thus, the distribution of the output of the causal projector follows $p(f_\phi(x_t)|x_t) \sim \mathcal{N}(\mathcal{Z},\delta^{2})$, and the log-likelihood of $p(f_\phi(x_t)|x_t)$ is expressed as:
\begin{small}
 \begin{equation}
    \log p(f_\phi(x_t)|x_t) = -0.5 \log(2\pi\delta^2) - (f_\phi(x_t) - \mathcal{Z})^2/2\delta^2
\end{equation}   
\end{small}
Thus, the Jacobian $J$ with regard to  $x_t$ can be expressed as:
\begin{small}
   \begin{equation}
\label{eq:causal-guidance}
    \nabla_{x_t} \log p(f_\phi(x_t)|x_t) = -(f_\phi(x_t) - \mathcal{Z})/\delta^2 \cdot \nabla_{x_t} f_\phi(x_t)
\end{equation} 
\end{small}
This gradient serves as the guidance which encourages the sampling process to generate images that align with the desired causal representation by updating the noisy image in the direction that maximizes the log-likelihood of the desired causal representation.

To adapt the previous sampling method to a deterministic sampling approach, such as denoising Diffusion Implicit Models (DDIM) \cite{ddim}, we can employ the score-based conditioning trick \cite{guided-diffusion,scored-based-diffusion}. With this deterministic sampling process, the generation at each timestep can be modified as follows:
\begin{small}
   \begin{equation}
\begin{split}
     x_{t-1} = & \sqrt{\bar\alpha_{t-1}} \left(\frac{x_t - \sqrt{1-\bar\alpha_t} (\epsilon_\theta(x_t, t) + \lambda \cdot (f_\phi(x_t) - \mathcal{Z})/\delta^2 \cdot \nabla_{x_t} f_\phi(x_t))}{\sqrt{\bar\alpha_t}}\right) \\ & + \sqrt{1-\bar\alpha_{t-1}} (\epsilon_\theta(x_t, t) + \lambda \cdot (f_\phi(x_t) - \mathcal{Z})/\delta^2\cdot \nabla_{x_t} f_\phi(x_t))
\end{split}
\end{equation} 
\end{small}
where $\epsilon_\theta(x_t, t)$ represents the predicted noise, $f_\phi(x_t)$ denotes the causal representation predicted from the noisy image $x_t$, $\mathcal{Z}$ is the desired generative factor for the counterfactual image, $\lambda$ is the guidance scale factor that controls the strength of the causal representation guidance, $\delta^2$ represents the variance of the assumed Gaussian distribution for the causal representation, and $\nabla_{x_t} f_\phi(x_t)$ is the gradient of the causal encoder network $f_\phi$ with respect to the noisy image $x_t$.
The guidance scale factor $\lambda$ controls the strength of the causal representation guidance. A higher value of $\lambda$ gives higher weight to the causal representation guidance on the generated image.

\begin{figure*}
    \centering
    \includegraphics[width=0.85\textwidth]{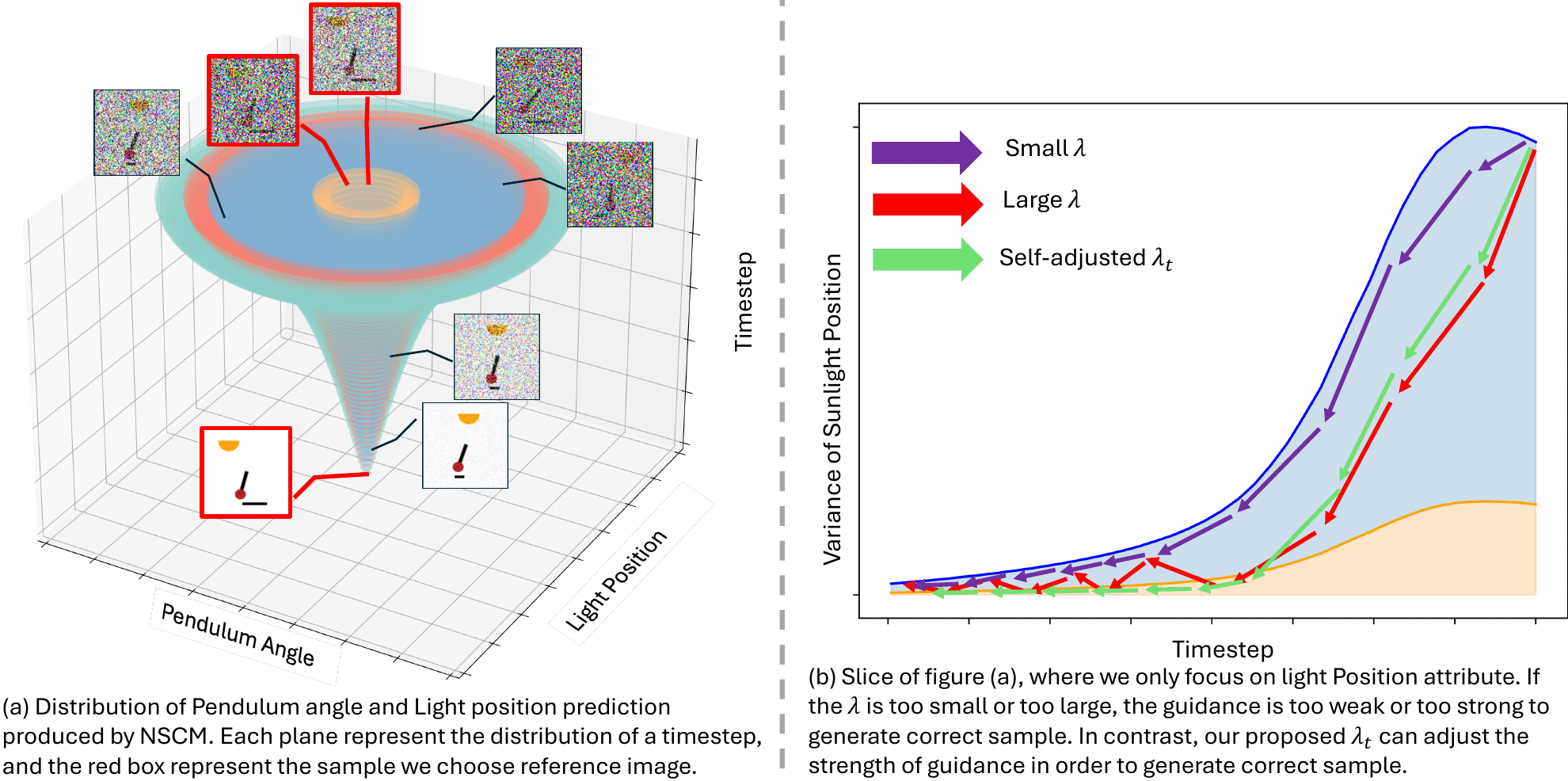}
    \caption{Illustration of the effects of different $\lambda$ values on guided sampling:
(a) A target sample (red box) is chosen, and the distribution of the difference in predicted generative factors (Pendulum angle and Light position) between the target sample and other samples is modeled as a 2D Gaussian using a causal predictor.
(b) Focusing on the Light position attribute, the orange area represents predictions from the noisy target sample, while the blue area represents predictions from other samples. Extreme $\lambda$ values fail to guide generation correctly. In contrast, our proposed self-adjusted $\lambda_t$ adjusts the strength for proper guidance.} %
    \label{fig:guidance-compare}
    \vspace{-\baselineskip}
\end{figure*}
\subsection{Self-Adjustment Scalar for Guidance}
\label{sec:adjusted-scalar}
In practice, to achieve good quality generation, we observe that the guidance scalar factor $\lambda$ not only needs to be carefully fine-tuned but  high quality generation is hard to be achieved with single $\lambda$ for all timesteps.  Recall that for each diffusion forward step, $x_t = \sqrt{\bar \alpha_t} x_0 + \sqrt{1-\bar \alpha_t}\epsilon$, where $\epsilon$ follows a standard normal distribution. As the timestep $t$ increases, the influence of the random noise $\epsilon$ becomes more dominant, causing the image $x_t$ to become more blurred. Therefore, as illustrated by \cref{fig:guidance-compare}, a larger scale factor $\lambda$ is needed when the timestep $t$ is large, \ie, earlier in the reverse sampling process.

To achieve this goal, we balance the variance of causal representation guidance term with the variance term in noise $x_t$. According to \cref{eq:causal-guidance}, the variance of causal representation guidance term can be expressed as:
\begin{small}
    \begin{equation}
    \text{Var}(\lambda_t \cdot \frac{(f_\phi(x_t) - \mathcal{Z})}{\delta^2} \cdot \nabla_{x_t} f_\phi(x_t)) = \lambda_t^2 \cdot \frac{\text{Var}(f_\phi(x_t))}{\delta^4} \cdot \|\nabla_{x_t} f_\phi(x_t)\|_2^2.
\end{equation}
\end{small}

We then apply the balance between causal representation guidance and uncertainty in the diffusion process by making the variance of causal representation guidance proportional to the variance of each diffusion timestep, which is controlled by the hyper-parameter $\gamma$:
\begin{small}
    \begin{equation}
\gamma^2 \cdot (1 - \bar\alpha_t) \cdot \sigma_t^2 = \lambda_t^2 \cdot \frac{\text{Var}(f_\phi(x_t))}{\delta^4} \cdot \|\nabla_{x_t} f_\phi(x_t)\|^2 = \lambda_t^2 \cdot \frac{1}{\delta^2} \cdot \|\nabla_{x_t} f_\phi(x_t)\|_2^2.
\end{equation}
\end{small}
Recall that $\epsilon$ for each diffusion timestep has identity variance, thus solving for $\lambda_t$, we obtain:
\begin{small}
  \begin{equation}
    \lambda_t = \sqrt{\frac{\gamma^2 \cdot (1 - \bar\alpha_t) \cdot \sigma_t^2 \cdot \delta^2}{\|\nabla_{x_t} f_\phi(x_t)\|_2^2}} = \sqrt{\frac{\gamma \cdot (1 - \bar\alpha_t) \cdot \delta^2}{\|\nabla_{x_t} f_\phi(x_t)\|_2^2}} = \gamma \cdot \frac{\sqrt{1-\bar \alpha_t}\delta}{\|\nabla_{x_t}f_{\phi}(x_t)\|_2}.
\end{equation}  
\end{small}

\section{Experiment Evaluation}
\label{sec:method}

\subsection{Benchmarks}
\label{sec:DataMetrics}

\textbf{Datasets ---}
We evaluate the effectiveness of the proposed framework for high-dimensional counterfactual outcomes generation on six datasets: Pendulum, Flow, CelebA(BEARD), CelebA(SMILE) \cite{DBLP:conf/cvpr/YangLCSHW21}, Shadow-Sunlight, and Shadow-Pointlight \cite{shadow}. Pendulum, Flow, CelebA(BEARD), and CelebA(SMILE) each contain four generative factors, while Shadow-Sunlight and Shadow-Pointlight are more challenging with seven and eight generative factors, respectively. These datasets cover diverse causal relations, from physical properties to facial attributes, providing a comprehensive assessment for the models

\textbf{Metrics ---} 
We use several standard metrics to evaluate the quality and diversity of the generated images. The Fréchet Inception Distance (FID) \cite{fid} uses a pre-trained Inception v3 model to measure the feature distribution differences between generated and real images, where lower FID scores indicate higher similarity. The Single Image FID (sFID) \cite{sFID} applies this concept to individual images by assessing them against the real image feature distribution. Peak Signal-to-Noise Ratio (PSNR) measures the ratio of the maximum possible signal power (the original image) to the noise power (the deviation from the original in the generated image), with higher values indicating better quality. 

While precision and recall \cite{PrecisionRecall} are widely used for evaluating categorically labeled generated images, they may not be suitable for counterfactual image models with continuous generative factors because precision and recall are designed for only predicting discrete, categorical labels.
This limitation can addressed by slightly modifying precision and recall, such that the underlying model is an attribute predictor, rather than a categorical classifier, which we call the Attribute Consistency Metric (ACM).
ACM is computed by calculating $L^2_2$ distance between the attribute values of the generated counterfactual samples, as predicted by an attribute predictor trained on the original dataset, and the desired attribute values.  The overall ACM score is obtained by averaging these distances across all generated images, with lower ACM scores indicating better performance. More details are in the Appendix.

\begin{table}
              \caption{Single counterfactual sample quality comparision with SOTA generative models for each tasks. Our method use DDIM deterministic sampler \cite{ddim}. All diffusion models are trained using 1000 steps. For CGD and our method, we use DDIM sampler with 100 steps. }
              \label{table:single-compare}
                \begin{minipage}{0.42\textwidth}
                \raggedright
                \renewcommand{\arraystretch}{1.2}
{
\begin{adjustbox}{width=\textwidth}
\begin{tabular}{lcccc}
\hlineB{2}
Model                             & FID $\downarrow$ & sFID$\downarrow$ & PSNR$\uparrow$ & ACM $\downarrow$ \\
\textbf{Pendulum (96$\times$ 96)} &       &    &  &          \\ \hline
DDPM \cite{ddpm}                            & 4.24       & 6.05    & 12.87 & 3.10    \\
CGD \cite{guided-diffusion}         & 3.96    &  5.95    &  12.76    & 3.01    \\
LDM  \cite{ldm}            & 3.93    &  5.57    &  12.97    & 2.96    \\
CausalVAE  \cite{DBLP:conf/cvpr/YangLCSHW21}    &9.69     &10.62      &7.65      & 3.67    \\
CausalGAN \cite{kocaoglu2018causalgan}           &6.89     &7.32      &11.48      &2.45     \\
CausalDiffusion      &\textbf{3.79}    & \textbf{4.92}     & \textbf{28.58}     & \textbf{1.39}    \\
                                  &     &      &      &     \\
\textbf{Flow (96$\times$ 96)}            &     &      &      &     \\ \hline
DDPM \cite{ddpm}                            & 3.02       & 4.17    & 20.87 & 2.10    \\
CGD \cite{guided-diffusion}         & 4.03    &  4.06    &  19.76    & 2.01    \\
LDM  \cite{ldm}            & 3.13    &  4.42    &  21.32    & 2.06    \\
CausalVAE  \cite{DBLP:conf/cvpr/YangLCSHW21}    &6.69     &7.62      &17.32      & 2.27    \\
CausalGAN \cite{kocaoglu2018causalgan}           &3.89     &3.92      &25.48      &1.97     \\
CausalDiffusion      &\textbf{2.87}    & \textbf{3.81}     & \textbf{31.14}     & \textbf{0.89}    \\
                                  &     &      &      &     \\
\textbf{CelebA(BEARD) (128$\times$ 128)} &     &      &      &     \\ \hline
DDPM \cite{ddpm}                            & 7.89       & 7.28    & N/A & 5.10    \\
CGD \cite{guided-diffusion}         & 7.02    &  6.86    &  N/A    & 5.21    \\
LDM  \cite{ldm}            & 6.92    &  \textbf{6.06}    &  N/A    & 4.26    \\
CausalVAE  \cite{DBLP:conf/cvpr/YangLCSHW21}    &32.03     &19.98      &N/A      & 6.67    \\
CausalGAN \cite{kocaoglu2018causalgan}           &8.14     &7.63      &N/A      &4.45     \\
CausalDiffusion      &\textbf{6.59}    & 6.09     & N/A     & \textbf{2.99}    \\\hlineB{2}
\end{tabular}
\end{adjustbox}
}
\end{minipage}
\hfill
\begin{minipage}[c]{0.43\textwidth}
\raggedright
 \renewcommand{\arraystretch}{1.2}
{
\begin{adjustbox}{width=\textwidth}
 \begin{tabular}{lcccc}
\hlineB{2}
Model                                 & FID $\downarrow$ & sFID$\downarrow$ & PSNR$\uparrow$ & ACM $\downarrow$ \\
\textbf{CelebA(SMILE) (128$\times$ 128)}     &     &      &      &     \\ \hline
DDPM \cite{ddpm}                            & 7.79       & 7.54    & N/A & 4.32    \\
CGD \cite{guided-diffusion}         & 7.21    &  7.04    &  N/A    & 4.03    \\
LDM  \cite{ldm}            & 7.03    &  7.22    &  N/A    & 4.06    \\
CausalVAE  \cite{DBLP:conf/cvpr/YangLCSHW21}    &31.49     &19.74      &N/A      & 5.67    \\
CausalGAN \cite{kocaoglu2018causalgan}           &8.09     &8.47      &N/A      &3.45     \\
CausalDiffusion      &\textbf{6.46}    & \textbf{6.72}     & N/A     & \textbf{2.61} \\
                                      &     &      &      &     \\
\textbf{Shadow-Sunlight (128$\times$ 128)}   &     &      &      &     \\ \hline
DDPM \cite{ddpm}                            & 5.02       & 8.12    & 11.76 & 6.20    \\
CGD \cite{guided-diffusion}         & 4.96    &  5.65    &  11.76    & 6.11    \\
LDM  \cite{ldm}            & 4.97    &  6.33    &  12.20    & 6.56    \\
CausalVAE  \cite{DBLP:conf/cvpr/YangLCSHW21}    &28.71     &19.38      &7.32      & 6.74    \\
CausalGAN \cite{kocaoglu2018causalgan}           &7.32     &6.04      &15.48      &3.44     \\
CausalDiffusion      &\textbf{4.61}    & \textbf{5.74}     & \textbf{21.14}     & \textbf{2.36}    \\
                                      &     &      &      &     \\
\textbf{Shadow-Pointlight (128$\times$ 128)} &     &      &      &     \\ \hline
DDPM \cite{ddpm}                            & 6.89       & 6.25    & 11.46 & 6.22    \\
CGD \cite{guided-diffusion}         & 6.96    &  6.65    &  11.68    & 6.35    \\
LDM  \cite{ldm}            & 6.93    &  6.57    &  12.77    & 5.97    \\
CausalVAE  \cite{DBLP:conf/cvpr/YangLCSHW21}    &29.77     &19.32      &9.65      & 5.69    \\
CausalGAN \cite{kocaoglu2018causalgan}           &8.99     &9.32      &15.32      &4.03     \\
CausalDiffusion      &\textbf{5.56}    & \textbf{5.94}     & \textbf{20.38}     & \textbf{2.79}    \\ \hlineB{2}
\end{tabular}
\end{adjustbox}
}
\end{minipage}%
\vspace{-0.2\baselineskip}
            \end{table}

\subsection{Comparing with SOTA Baselines}
\label{sec:compare}
We compare the proposed framework against SOTA for counterfactual image generation, including CausalVAE and CausalGAN \cite{kocaoglu2018causalgan}, which incorporates causal knowledge into generation. To evaluate the impact of incorporating causal knowledge and compare the performance of diffusion-based models in inferring counterfactual outcomes, we also include Denoising Diffusion Probabilistic Models (DDPM) \cite{ddpm}, Classifier-Guided Diffusion (CGD) \cite{guided-diffusion}, and Latent Diffusion Models (LDM) \cite{ldm} in our experiments.
\begin{figure*}
    \centering
     \includegraphics[width=0.93\textwidth]{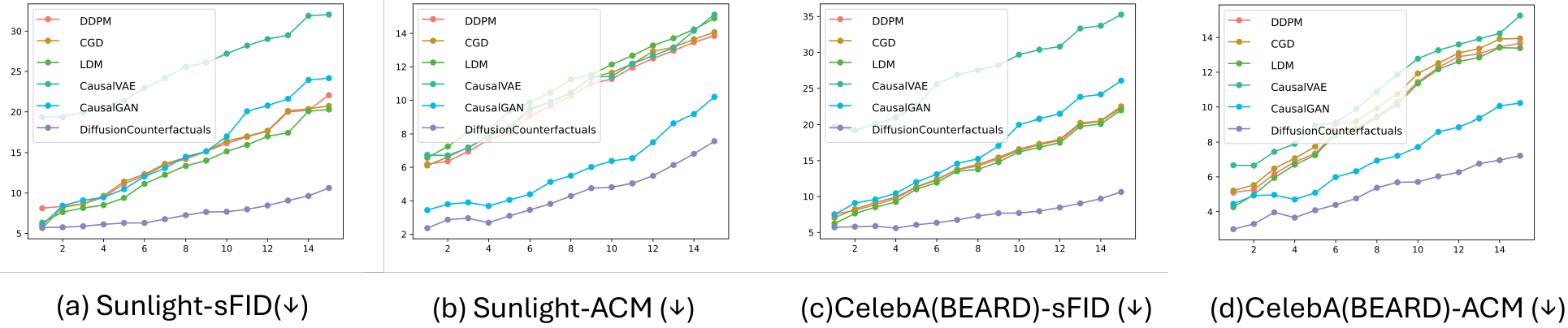}
    \caption{Sequential counterfactual generations results. We evaluate the sequential counterfactual generated samples in an autoregressive way.  More results are included in Appendix.}
    \label{fig:squential}
    \vspace{-1\baselineskip}
\end{figure*}

\textbf{Single Counterfactual Sample Generation Comparison:} The proposed method demonstrates superior performance compared to other approaches using standard metrics, including FID, sFID, and PSNR, with only a 0.03 sFID score difference from the best-performing method on CelebA(BEARD), as shown in  \cref{table:single-compare}.
 By discovering the underlying causal mechanisms and utilizing them to infer counterfactual outcomes, our method also achieves the best performance on the ACM metric. Although CausalVAE and CausalGAN can infer counterfactual outcomes in the generative factors space, their limited generation abilities prevent them from producing high-quality counterfactual outcomes in the input space. This limitation compromises their performance on the ACM metrics.

\textbf{Sequential Counterfactual Sample Generation Comparison:} As discussed in \cref{sec:introduction}, incorporating causality into the generation model improves the consistency of samples in a sequence. We evaluate this by assessing the counterfactual samples in a step-by-step autoregressive manner, where each generated sample serves as the input for producing the subsequent sample. For instance, in the Pendulum dataset, we sequentially adjust the light position and generate each subsequent counterfactual sample based on the updated light position. For more details, please refer to  the Appendix. As demonstrated in \cref{fig:squential},  our method consistently generates high-quality counterfactual samples across multiple steps. In contrast, baseline methods that do not incorporate causal knowledge struggle to maintain consistency among the generated samples. This shows the importance of using causal knowledge in the generation process to get coherent and accurate results.

\begin{table}
\caption{Compare with Simple integration approaches.}
\label{table:simple-integration}
                \begin{minipage}{0.43\textwidth}
                \raggedright
                \renewcommand{\arraystretch}{1.2}
{
\begin{adjustbox}{width=\textwidth}
\begin{tabular}{lcccc}
\hline
Model                             & FID $\downarrow$ & sFID$\downarrow$ & PSNR$\uparrow$ & ACM $\downarrow$                \\
\textbf{Pendulum (96$\times$ 96)}         &                      &                      &                      &                      \\ \hline
LDM+NSCM                        &  4.21                    & 5.96                     &  16.68                    &  2.96        \\
CF+NSCM                           & 4.32 & 6.02 & 17.86 & 2.77 \\
CausalDiffusion      &\textbf{3.79}    & \textbf{4.92}     & \textbf{28.58}     & \textbf{0.99}        \\
                                  &                      &                      &                      &                      \\
\textbf{Flow (96$\times$ 96)}            &                      &                      &                      &                               \\ \hline
LDM+NSCM                         &  3.11                    & 4.16                     &   24.65                   &  1.97                    \\
CF+NSCM                           &  3.06                    & 4.09                     & 25.67                     &  1.64 {} \\
CausalDiffusion      &\textbf{2.87}    & \textbf{3.81}     & \textbf{31.14}     & \textbf{0.89}    \\
                                  &                      &                      &                      &                      \\
\textbf{CelebA(BEARD) (128$\times$ 128)} &                      &                      &                      &                      \\ \hline
LDM+NSCM                          &  6.96                    & 8.08                     &   N/A                   &  4.06                    \\
CF+NSCM                           & 6.97 & 7.91 & N/A & 3.75 \\
CausalDiffusion      &\textbf{6.59}    & 6.09     & N/A     & \textbf{2.99}                  \\ \hline
\end{tabular}
\end{adjustbox}
}
\end{minipage}
\hfill
\begin{minipage}[c]{0.43\textwidth}
\raggedright
 \renewcommand{\arraystretch}{1.2}
{
\begin{adjustbox}{width=\textwidth}
\begin{tabular}{lcccc}
\hline
Model                                & FID $\downarrow$ & sFID$\downarrow$ & PSNR$\uparrow$ & ACM $\downarrow$                  \\
\textbf{CelebA(SMILE) (128$\times$ 128)}     &                     &                    &                    &                      \\ \hline
LDM+NSCM                              &  6.77                    &   7.06                   &   N/A                   & 3.67                     \\
CF+NSCM                               & 6.93 & 7.04 & N/A & 3.32 \\
CausalDiffusion      &\textbf{6.46}    & \textbf{6.72}     & N/A     & \textbf{2.61}                     \\
                                      &                      &                      &                      &                      \\
\textbf{Shadow-Sunlight (128$\times$ 128)}   &                      &                      &                      &                      \\ \hline
LDM+NSCM                              & 4.98                &  7.17                    & 16.68                     &  3.97                    \\
CF+NSCM                               & 5.02 & 6.18 & 16.94 & 3.67 \\
CausalDiffusion      &\textbf{4.61}    & \textbf{5.74}     & \textbf{21.14}     & \textbf{2.36}                    \\
                                      &                      &                      &                      &                      \\
\textbf{Shadow-Pointlight (128$\times$ 128)} &                      &                      &                      &                      \\ \hline
LDM+NSCM                              &   6.26                   &  7.15                    & 15.59                     &  3.97                    \\
CF+NSCM                               & 5.92 & \textbf{5.88} & 17.76 & 3.91 \\
CausalDiffusion      &\textbf{5.56}    & 5.94     & \textbf{20.38}     & \textbf{2.79}    \\ \hlineB{2}
\end{tabular}
\end{adjustbox}
}
\end{minipage}%
\vspace{-\baselineskip}
\end{table}

\begin{table}
              \label{table:compare-building-block}
                \begin{minipage}{0.44\textwidth}
                \caption{Ablation study of comparing different loss and the weights.}
                \label{table:ablation-train}
                \renewcommand{\arraystretch}{1.2}
{
\begin{adjustbox}{width=\textwidth}
\begin{tabular}{lcccc}
\hline
Model                              & FID $\downarrow$ & sFID$\downarrow$ & PSNR$\uparrow$ & ACM $\downarrow$                 \\
\textbf{CelebA(BEARD) (128$\times$ 128)}     &                      &                      &                      &                      \\ \hline
w/o contrastive                      & 6.76                      &   6.12                   &   N/A                   &   3.27                   \\
w/o w(t)                             & 6.89 & 6.69 & N/A & 3.19 \\
the proposed (ours)                 &\textbf{6.59}    & 6.09     & N/A     & \textbf{2.99}                   \\
                                     & \multicolumn{1}{l}{} & \multicolumn{1}{l}{} & \multicolumn{1}{l}{} & \multicolumn{1}{l}{} \\
\textbf{Shadow-Pointlight(128$\times$ 128)} &                      &                      &                      &                      \\ \hline
w/o contrastive                      &  5.97                    & 6.02                     &   17.76                   & 3.06                     \\
w/o w(t)                             & 6.02 & 6.83 & 17.91& 3.11 \\
the proposed (ours)             &\textbf{5.56}    & 5.94     & \textbf{20.38}     & \textbf{2.79}                    \\ \hlineB{2}
\end{tabular}
\end{adjustbox}
}
\end{minipage}
\hfill
\begin{minipage}[c]{0.45\textwidth}
\caption{Ablation study of comparing different $\lambda$.}
                \label{table:ablation-sampling}
\raggedright
 \renewcommand{\arraystretch}{1.2}
{
\begin{adjustbox}{width=\textwidth}
\begin{tabular}{lcccc}
\hline
Model                                & FID $\downarrow$ & sFID$\downarrow$ & PSNR$\uparrow$ & ACM $\downarrow$                  \\
\textbf{CelebA(BEARD) (128$\times$ 128)}     &                      &                      &                      &                      \\ \hline
Fixed lambda                         & 7.29                      & 7.06                     &  N/A                    & 3.74                     \\
Trainable lambda                     & 6.95 & 7.26 & N/A & 3.66 \\
Linear lambda                        & 7.01                     & 7.24                     &   N/A                   & 3.70                    \\
Self-Adjusted (ours)                &\textbf{6.59}    & 6.09     & N/A     & \textbf{2.99}                  \\
                                     & \multicolumn{1}{l}{} & \multicolumn{1}{l}{} & \multicolumn{1}{l}{} & \multicolumn{1}{l}{} \\
\textbf{Shadow-Pointlight(128$\times$ 128)} &                      &                      &                      &                      \\ \hline
Fixed lambda                       &  5.16                    &   6.97                   & 16.61
& 3.49                     \\
Trainable lambda                     & 5.17 & 6.22 & 16.98 & 3.43 \\
Linear lambda                        &  5.32                    &  6.71                    &  15.76                    & 4.03                     \\
Self-Adjusted (ours)             &\textbf{5.56}    & 5.94     & \textbf{20.38}     & \textbf{2.79}                    \\ \hlineB{2}
\end{tabular}
\end{adjustbox}
}
\end{minipage}%
\vspace{-\baselineskip}
            \end{table}
\subsection{Discussion and Ablation Study}
\textbf{Comparison with Simple Integration Approaches:} In this section, we compare the proposed method with two simple integration approaches --- (1) incorporating causality into diffusion models as independent components and (2) using causal representations for classifier-free method \cite{diffcf}. The first approach involves a straightforward combination of causality and diffusion models, where a causal projector generates a counterfactual representation that is fed into the Latent Diffusion Model (LDM) as a conditioning input for generation, similar to independent components. The second approach uses causal representations for classifier-free generation. 
As shown in \cref{table:simple-integration},  DiffusionCounterfactuals  outperforms simple integration approaches.
The suboptimal performance of these simple combinations can be attributed to two main issues. Firstly, they overlook the prior inductive bias, which assumes that the residual between predicted and target values follows a normal distribution. Ignoring this assumption renders the counterfactual generation guidance unreliable. Secondly, these combinations lack the time-dependent control featured in our method, which is crucial for modulating the guidance strength across different timesteps. The absence of this control leads to unstable guidance and suboptimal outcomes.

\textbf{Conditional Timestep Training and Loss Function:} Our proposed method aligns high-dimensional raw inputs with causal generative factors during training, employing mean squared error (MSE) loss for feature-level alignment and contrastive loss for logit-level alignment. Additionally, we introduce conditional timestep weights to enhance both types of loss and improve counterfactual inference. These weights modulate the loss based on the diffusion timestep, allowing the model to adapt its training dynamics according to the level of noise present in the input. As demonstrated in \cref{table:ablation-train}, removing either the contrastive loss or the conditional timestep weights reduces model performance.

\textbf{Discussion on Self-adjusted Scalar for guidance:} During the counterfactual inference stage, a self-adjusted scalar is utilized to control the counterfactual causal condition strength. To demonstrate the effectiveness of the self-adjusted scalar, we compare it with three alternative approaches --- (1) fixed scalar $\lambda$, (2) trainable but time-invariant scalar $\lambda_{tr}$, and (3) simple linear decay scalar $\lambda_{linear}$. As shown in \cref{table:ablation-sampling}, fixed $\lambda$ and trainable $\lambda_{tr}$ can not achieve satisfactory performance. Although the simple linear decay approach ($\lambda_{linear}$) improves performance, it still yields sub-optimal results compared to the self-adjusted scalar utilized in the proposed method. 
As illustrated in \cref{fig:guidance-compare}, the guidance strength for each timestep is nonlinear. Therefore,
$\lambda_{linear}$ still achieves sub-optimal performance compared to the self-adjusted scalar utilized in proposed method.

\vspace{-0.2\baselineskip}
\section{Conclusion}
\vspace{-0.3\baselineskip}
In this paper, we propose DiffusionCounterfactuals, a framework integrating diffusion models with causal representation learning to generate accurate and consistent counterfactual outcomes. DiffusionCounterfactuals uses a modified sampling process with gradient-based guidance and a self-adjusting scalar. Experiments on six datasets along with detail discussions and ablatiosn show the superiority of DiffusionCounterfactuals in generating high-quality, causally consistent counterfactuals.

\clearpage

\bibliographystyle{ieee_fullname}
\bibliography{egbib}

\begin{thebibliography}{10}\itemsep=-1pt

\bibitem{counterfactual-policy}
Onur Atan, William~R. Zame, Qiaojun Feng, and Mihaela van~der Schaar.
\newblock Constructing effective personalized policies using counterfactual inference from biased data sets with many features.
\newblock {\em Machine Learning}, 108(6):945--970, Jun 2019.

\bibitem{simclr}
Ting Chen, Simon Kornblith, Mohammad Norouzi, and Geoffrey Hinton.
\newblock A simple framework for contrastive learning of visual representations.
\newblock In Hal~Daumé III and Aarti Singh, editors, {\em Proceedings of the 37th International Conference on Machine Learning}, volume 119 of {\em Proceedings of Machine Learning Research}, pages 1597--1607. PMLR, 13--18 Jul 2020.

\bibitem{counterfactual-economics}
Victor Chernozhukov, Iván Fernández-Val, and Blaise Melly.
\newblock Inference on counterfactual distributions.
\newblock {\em Econometrica}, 81(6):2205--2268, 2013.

\bibitem{guided-diffusion}
Prafulla Dhariwal and Alexander Nichol.
\newblock Diffusion models beat gans on image synthesis.
\newblock In M. Ranzato, A. Beygelzimer, Y. Dauphin, P.S. Liang, and J.~Wortman Vaughan, editors, {\em Advances in Neural Information Processing Systems}, volume~34, pages 8780--8794. Curran Associates, Inc., 2021.

\bibitem{deci}
Tomas Geffner, Javier Antoran, Adam Foster, Wenbo Gong, Chao Ma, Emre Kiciman, Amit Sharma, Angus Lamb, Martin Kukla, Nick Pawlowski, Miltiadis Allamanis, and Cheng Zhang.
\newblock Deep end-to-end causal inference.
\newblock In {\em NeurIPS 2022 Workshop on Causality for Real-world Impact}, 2022.

\bibitem{gan}
Ian Goodfellow, Jean Pouget-Abadie, Mehdi Mirza, Bing Xu, David Warde-Farley, Sherjil Ozair, Aaron Courville, and Yoshua Bengio.
\newblock Generative adversarial nets.
\newblock In Z. Ghahramani, M. Welling, C. Cortes, N. Lawrence, and K.Q. Weinberger, editors, {\em Advances in Neural Information Processing Systems}, volume~27. Curran Associates, Inc., 2014.

\bibitem{fid}
Martin Heusel, Hubert Ramsauer, Thomas Unterthiner, Bernhard Nessler, and Sepp Hochreiter.
\newblock Gans trained by a two time-scale update rule converge to a local nash equilibrium.
\newblock In I. Guyon, U.~Von Luxburg, S. Bengio, H. Wallach, R. Fergus, S. Vishwanathan, and R. Garnett, editors, {\em Advances in Neural Information Processing Systems}, volume~30. Curran Associates, Inc., 2017.

\bibitem{DBLP:conf/iclr/HigginsMPBGBML17}
Irina Higgins, Lo{\"{\i}}c Matthey, Arka Pal, Christopher Burgess, Xavier Glorot, Matthew Botvinick, Shakir Mohamed, and Alexander Lerchner.
\newblock beta-vae: Learning basic visual concepts with a constrained variational framework.
\newblock In {\em 5th International Conference on Learning Representations, {ICLR} 2017, Toulon, France, April 24-26, 2017, Conference Track Proceedings}. OpenReview.net, 2017.

\bibitem{ddpm}
Jonathan Ho, Ajay Jain, and Pieter Abbeel.
\newblock Denoising diffusion probabilistic models.
\newblock In H. Larochelle, M. Ranzato, R. Hadsell, M.F. Balcan, and H. Lin, editors, {\em Advances in Neural Information Processing Systems}, volume~33, pages 6840--6851. Curran Associates, Inc., 2020.

\bibitem{diffusion-explain}
Guillaume Jeanneret, Loïc Simon, and Frédéric Jurie.
\newblock Diffusion models for counterfactual explanations.
\newblock In {\em ACCV}, 2022.

\bibitem{avoid-discrimination}
Niki Kilbertus, Mateo Rojas~Carulla, Giambattista Parascandolo, Moritz Hardt, Dominik Janzing, and Bernhard Sch\"{o}lkopf.
\newblock Avoiding discrimination through causal reasoning.
\newblock In I. Guyon, U.~Von Luxburg, S. Bengio, H. Wallach, R. Fergus, S. Vishwanathan, and R. Garnett, editors, {\em Advances in Neural Information Processing Systems}, volume~30. Curran Associates, Inc., 2017.

\bibitem{kocaoglu2018causalgan}
Murat Kocaoglu, Christopher Snyder, Alexandros~G. Dimakis, and Sriram Vishwanath.
\newblock Causal{GAN}: Learning causal implicit generative models with adversarial training.
\newblock In {\em International Conference on Learning Representations}, 2018.

\bibitem{diffusion-audio}
Zhifeng Kong, Wei Ping, Jiaji Huang, Kexin Zhao, and Bryan Catanzaro.
\newblock Diffwave: A versatile diffusion model for audio synthesis.
\newblock In {\em International Conference on Learning Representations}, 2021.

\bibitem{PrecisionRecall}
Tuomas Kynkäänniemi, Tero Karras, Samuli Laine, Jaakko Lehtinen, and Timo Aila.
\newblock Improved precision and recall metric for assessing generative models.
\newblock {\em CoRR}, abs/1904.06991, 2019.

\bibitem{social_cf}
Jack~S. Levy.
\newblock Counterfactuals, causal inference, and historical analysis.
\newblock {\em Security Studies}, 24(3):378--402, 2015.

\bibitem{diffusion-text}
Xiang~Lisa Li, John Thickstun, Ishaan Gulrajani, Percy Liang, and Tatsunori Hashimoto.
\newblock Diffusion-{LM} improves controllable text generation.
\newblock In Alice~H. Oh, Alekh Agarwal, Danielle Belgrave, and Kyunghyun Cho, editors, {\em Advances in Neural Information Processing Systems}, 2022.

\bibitem{sFID}
Charlie Nash, Jacob Menick, Sander Dieleman, and Peter~W. Battaglia.
\newblock Generating images with sparse representations.
\newblock {\em ArXiv}, abs/2103.03841, 2021.

\bibitem{bib:gae}
Ignavier Ng, Shengyu Zhu, Zhitang Chen, and Zhuangyan Fang.
\newblock A graph autoencoder approach to causal structure learning.
\newblock {\em arXiv preprint arXiv:1911.07420}, 2019.

\bibitem{Nichol2021ImprovedModels}
Alex Nichol and Prafulla Dhariwal.
\newblock {Improved Denoising Diffusion Probabilistic Models}.
\newblock {\em arxiv pre-print}, 12 2021.

\bibitem{diffcf}
Alex Nichol, Prafulla Dhariwal, Aditya Ramesh, Pranav Shyam, Pamela Mishkin, Bob McGrew, Ilya Sutskever, and Mark Chen.
\newblock Glide: Towards photorealistic image generation and editing with text-guided diffusion models.
\newblock In {\em Proceedings of the 39th International Conference on Machine}, 2022.

\bibitem{reason:Pearl09a}
Judea Pearl.
\newblock {\em Causality: Models, Reasoning and Inference}.
\newblock Cambridge University Press, 2nd edition, 2009.

\bibitem{counterfactual-medicine}
Jonathan~G. Richens, Ciar{\'a}n~M. Lee, and Saurabh Johri.
\newblock Improving the accuracy of medical diagnosis with causal machine learning.
\newblock {\em Nature Communications}, 11(1):3923, Aug 2020.

\bibitem{medical_cf}
Jonathan~G. Richens, Ciar{\'a}n~M. Lee, and Saurabh Johri.
\newblock Improving the accuracy of medical diagnosis with causal machine learning.
\newblock {\em Nature Communications}, 11(1):3923, Aug 2020.

\bibitem{ldm}
Robin Rombach, Andreas Blattmann, Dominik Lorenz, Patrick Esser, and Björn Ommer.
\newblock High-resolution image synthesis with latent diffusion models, 2021.

\bibitem{unet}
O Ronneberger, {P.Fischer}, and T Brox.
\newblock {U-Net: Convolutional Networks for Biomedical Image Segmentation}.
\newblock In {\em Proc. of Medical Image Computing and Computer-Assisted Intervention}, volume 9351, pages 234--241. Springer, 2015.

\bibitem{diffscm}
Pedro Sanchez and Sotirios~A. Tsaftaris.
\newblock Diffusion causal models for counterfactual estimation.
\newblock In {\em First Conference on Causal Learning and Reasoning}, 2022.

\bibitem{bib:towards-causal}
Bernhard Schölkopf, Francesco Locatello, Stefan Bauer, Nan~Rosemary Ke, Nal Kalchbrenner, Anirudh Goyal, and Yoshua Bengio.
\newblock Toward causal representation learning.
\newblock {\em Proceedings of the IEEE}, 109(5):612--634, 2021.

\bibitem{first-diffusion}
Jascha Sohl-Dickstein, Eric Weiss, Niru Maheswaranathan, and Surya Ganguli.
\newblock Deep unsupervised learning using nonequilibrium thermodynamics.
\newblock In Francis Bach and David Blei, editors, {\em Proceedings of the 32nd International Conference on Machine Learning}, volume~37 of {\em Proceedings of Machine Learning Research}, pages 2256--2265, Lille, France, 07--09 Jul 2015. PMLR.

\bibitem{ddim}
Jiaming Song, Chenlin Meng, and Stefano Ermon.
\newblock Denoising diffusion implicit models.
\newblock In {\em International Conference on Learning Representations}, 2021.

\bibitem{scored-based-diffusion}
Yang Song, Jascha Sohl-Dickstein, Diederik~P Kingma, Abhishek Kumar, Stefano Ermon, and Ben Poole.
\newblock Score-based generative modeling through stochastic differential equations.
\newblock In {\em International Conference on Learning Representations}, 2021.

\bibitem{Vaswani2017AttentionNeed}
Ashish Vaswani, Noam Shazeer, Niki Parmar, Jakob Uszkoreit, Llion Jones, Aidan~N Gomez, Łukasz Kaiser, and Illia Polosukhin.
\newblock {Attention Is All You Need}.
\newblock In {\em Advances in neural information processing systems}, pages 5998--6008, 2017.

\bibitem{DBLP:conf/cvpr/YangLCSHW21}
Mengyue Yang, Furui Liu, Zhitang Chen, Xinwei Shen, Jianye Hao, and Jun Wang.
\newblock Causalvae: Disentangled representation learning via neural structural causal models.
\newblock In {\em {IEEE} Conference on Computer Vision and Pattern Recognition, {CVPR} 2021, virtual, June 19-25, 2021}, pages 9593--9602. Computer Vision Foundation / {IEEE}, 2021.

\bibitem{bib:notears}
Xun Zheng, Bryon Aragam, Pradeep Ravikumar, and Eric~P. Xing.
\newblock {DAGs with NO TEARS: Continuous Optimization for Structure Learning}.
\newblock In {\em Advances in Neural Information Processing Systems}, 2018.

\bibitem{bib:do-VAE}
Jiageng Zhu, Hanchen Xie, and Wael AbdAlmgaeed.
\newblock Do-operation guided causal representation learning with reduced supervision strength.
\newblock In {\em NeurIPS 2022 Workshop on Causality for Real-world Impact}, 2022.

\bibitem{shadow}
Jiageng Zhu, Hanchen Xie, Jianhua Wu, Jiazhi Li, Mahyar Khayatkhoei, Mohamed~E. Hussein, and Wael AbdAlmageed.
\newblock Shadow datasets, new challenging datasets for causal representation learning, 2023.

\end{thebibliography}

\clearpage
\newpage

\newpage 
\appendix
\section{Detail Discussion About DiffusionCounterfactual}
In \cref{sec:causal-diffusion}, we describe the problem setup, the training objective with practical loss and the reverse sampling. In this section, we mainly discuss the training objective in theoretical with more detail. 

As mentioned in \cref{sec:causal-diffusion}, for counterfactual inference in input space, the overall goal is to approximate $p(\hat{x}|x, I)$, where $x$ is the original input, $I$ is the intervention and the $\hat{x}$ is the expected counterfactual inference in input space. In our setting, we use causal projector $h_\phi$ to approximate $p(z|x)$ and the neural structural causal model (NSCM) $s_\phi$ to approximate 

To derive the Evidence Lower Bound (ELBO) for your specific setup, where you have the following model components:
- \( h_{\phi}(x) \) as the variational approximation of \( p(z|x) \)
- \( s_{\phi}(z, I) \) as the variational approximation of \( p(\hat{z}|z, I) \)
- \( g_{\theta}(\hat{z}) \) as the approximation of \( p(\hat{x}|\hat{z}) \)

The goal is to maximize the log likelihood \( \log p(\hat{x}|x, I) \) which can be written as:

\begin{equation}
    \log p(\hat{x}|x, I) = \log \int \int p(\hat{x}, \hat{z}, z | x, I) \, d\hat{z} \, dz 
\end{equation}

This integral is typically intractable, so we approximate it using variational inference. We introduce variational distributions \( h_{\phi}(z|x) \) and \( s_{\phi}(\hat{z}|z, I) \) to estimate the posterior distributions \( p(z|x) \) and \( p(\hat{z}|z, I) \), respectively. The approximation approach uses the following identity based on the properties of logarithms and expectations:

\begin{equation}
    \log p(\hat{x}|x, I) = \log \int \frac{p(\hat{x}, \hat{z}, z | x, I)}{q(\hat{z}, z | x, I)} q(\hat{z}, z | x, I) \, d\hat{z} \, dz
\end{equation}

Using Jensen's Inequality, we obtain:

\begin{equation}
     \log p(\hat{x}|x, I) \geq \int q(\hat{z}, z | x, I) \log \frac{p(\hat{x}, \hat{z}, z | x, I)}{q(\hat{z}, z | x, I)} \, d\hat{z} \, dz
\end{equation}

This inequality introduces the ELBO, which is the right-hand side of the inequality. Expanding the terms:

\begin{equation}
    \begin{split}
        p(\hat{x}, \hat{z}, z | x, I) = p(\hat{x} | \hat{z}) p(\hat{z} | z, I) p(z | x) \\
        q(\hat{z}, z | x, I) = s_{\phi}(\hat{z} | z, I) h_{\phi}(z | x) 
    \end{split}
\end{equation}

Substituting these into the ELBO, we get:

\begin{equation}
    \text{ELBO} = \mathbb{E}_{q(\hat{z}, z | x, I)} \left[ \log \frac{p(\hat{x} | \hat{z}) p(\hat{z} | z, I) p(z | x)}{s_{\phi}(\hat{z} | z, I) h_{\phi}(z | x)} \right]
\end{equation}

Splitting the log terms and rearranging gives:
\begin{equation}
     \text{ELBO} = \mathbb{E}_{h_{\phi}(z|x) s_{\phi}(\hat{z}|z, I)} \left[ \log p(\hat{x}|\hat{z}) \right] - \text{KL}(h_{\phi}(z|x) \| p(z|x)) -
\text{KL}(s_{\phi}(\hat{z}|z, I) \| p(\hat{z}|z, I)) 
\end{equation}

Where the first term is the expected log likelihood of the generated \(\hat{x}\) given the latent variable \(\hat{z}\), often interpreted as the reconstruction loss.
The second and third terms are the Kullback-Leibler divergence between the variational distributions and the true posterior distributions, serving as regularization terms that enforce the learned distributions to be close to the true distributions.

By maximizing the ELBO, we indirectly maximize the log likelihood of observing \(\hat{x}\) given \(x\) and \(I\), while keeping the variational distributions \( h_{\phi}(z|x) \) and \( s_{\phi}(\hat{z}|z, I) \) close to the true posterior distributions. 

Since one sample in the dataset is actually counterfactual of another sample, i.e., intervening on the root cause factors, we can use $\log p(x|z)$ to replace $\log p(\hat{x}|\hat{z})$, which is regular diffusion generation with condition. For the second term, we can optimize this term using MSE loss in the generative factor space. For the third term, the goal is to learn the intervention outcome in the generative factor space, which can be optimized by learned the correct causal graph and the corresponding causal mechanism \cite{reason:Pearl09a,deci,bib:gae}. Therefore, the third term can be optimized by discovering correct causal mechanism \cite{deci}, which can be expressed as following:

\begin{equation}
    min~\text{KL}(s_{\phi}(\hat{z}|z, I) \| p(\hat{z}|z, I)) \Leftrightarrow max~\mathbb{E}_{q_{s_\phi, \mathcal{G}}} \log p(G)\prod_{n} p_\phi(Z^{(i)}|G)+ \log p_{s_\phi}(z'|x,z)
\end{equation}

By the more theoretical analysis, we can further justify the model design and loss choices discussed in \cref{sec:causal-diffusion}.

\section{Metrics Details}
\subsection{ACM Metrics Details}
\begin{figure*}
    \centering
     \includegraphics[width=0.93\textwidth]{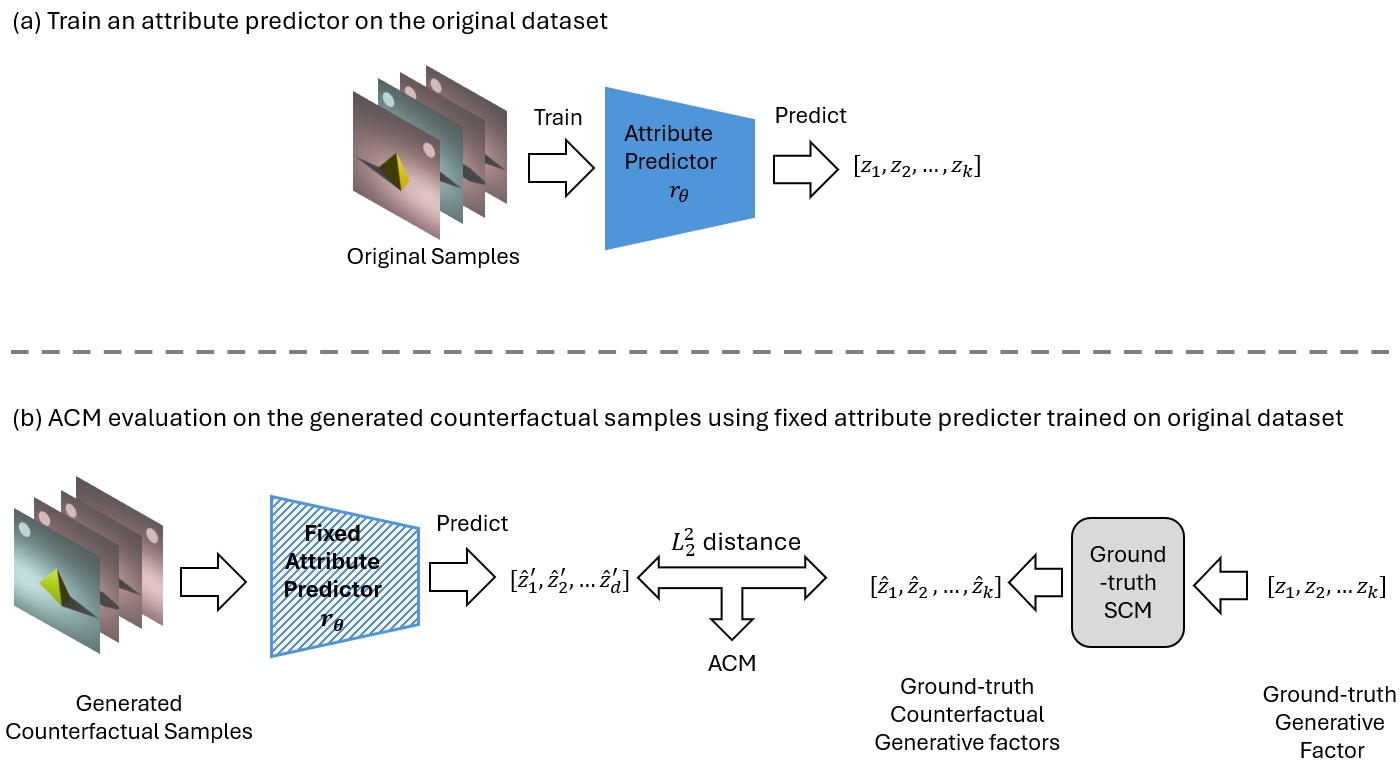}
    \caption{Attribute Consistency Metrics (ACM) calculation pipeline.}
    \label{fig:acm}
    \vspace{-\baselineskip}
\end{figure*}

As discussed in \cref{sec:DataMetrics}, Precision and Recall \cite{PrecisionRecall} was proposed to evaluate the quality of generated image when the image can be described by a class label (e.g., a dog or a cat). However, in our case, samples in a dataset are generative by sets of multiple continuous generative factors, which we can not use classification metrics to predict a sample in the generated images. Therefore, we propose attribute consistency metric (ACM) to properly evaluate the quality of image generation in this case. Similar with the Precision and Recall, we first train predictor using samples in the original datasets and their corresponding ground-truth generative factors. Once such predictor is well trained, we use it to make predictions on the generated images. After that, the $L^2_2$ distance between predicted attributes and the ground-truth generative factors is calculated. The whole ACM evaluation pipeline can be illustrated in \cref{fig:acm} 

According to the \cref{fig:acm}, we first train an attribute predictor $r_{\theta}$ on the original dataset using original images and their corresponding generative factors. Then the ACM for the generated counterfactual images can be evaluated as:
\begin{equation}
    ACM = \frac{1}{N} \sum_i^N || r_\theta(\hat{x}'^{(i)}) - f(\mathcal{Z},I) ||_2^2
\end{equation}

where $\hat{x}'^{(i)}$ is a generated counterfactual image, $f$ is ground-truth structural causal model and $I$ is intervention applied to generate ground-truth counterfactual generative factors.

\subsection{Other Metrics Details}
For FID \cite{fid}, sFID \cite{sFID} and PSNR metrics, to ensure consistent comparisons, we use the entire training set as the reference batch and evaluate metrics for all models using the same codebase.  For synthetic datasets including Pendulum, Flow \cite{DBLP:conf/cvpr/YangLCSHW21}, Shadow-Sunlight and Shadow-Pointlight \cite{shadow}, the PSNR can be measured for the generated counterfactual images. Because the original papers provide the code/project to generate ground-truth counterfactual images. However, for the real datasets, such as CelebA(BEARD) and CelebA(SMILE), PSNR can not be calculated for the generated counterfactual images because there is no ground-truth counterfactual image exist, for example, how a person looks like when becoming older. Therefore, for PSNR term in \cref{table:single-compare,table:simple-integration,table:ablation-train,table:ablation-sampling} are not calculated.

\section{More Experiment results}
\begin{figure*}
    \centering
     \includegraphics[width=0.93\textwidth]{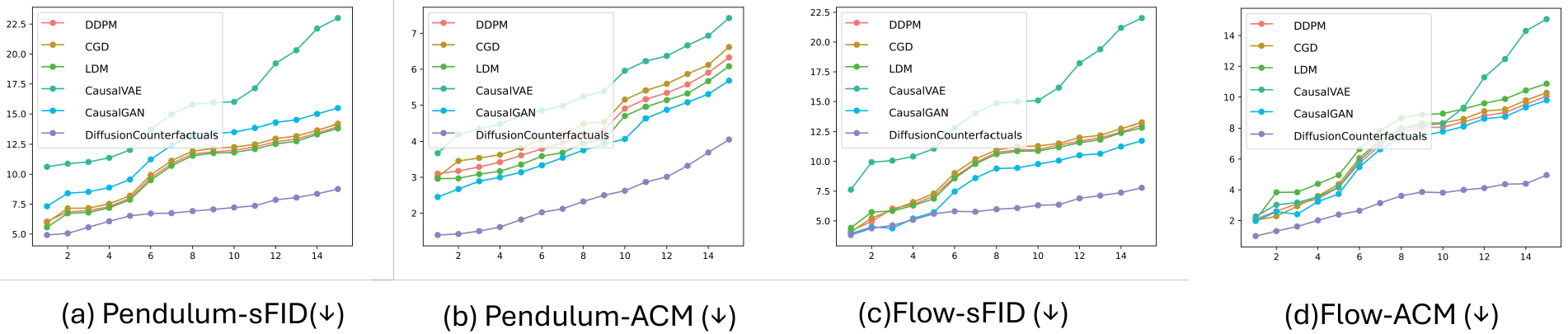}
    \caption{Sequential counterfactual generations results on Pendulum and Flow dataset. We evaluate the sequential counterfactual generated samples in an autoregressive way.}
    \label{fig:squential2}
    \vspace{-\baselineskip}
\end{figure*}

\begin{figure*}
    \centering
\includegraphics[width=0.93\textwidth]{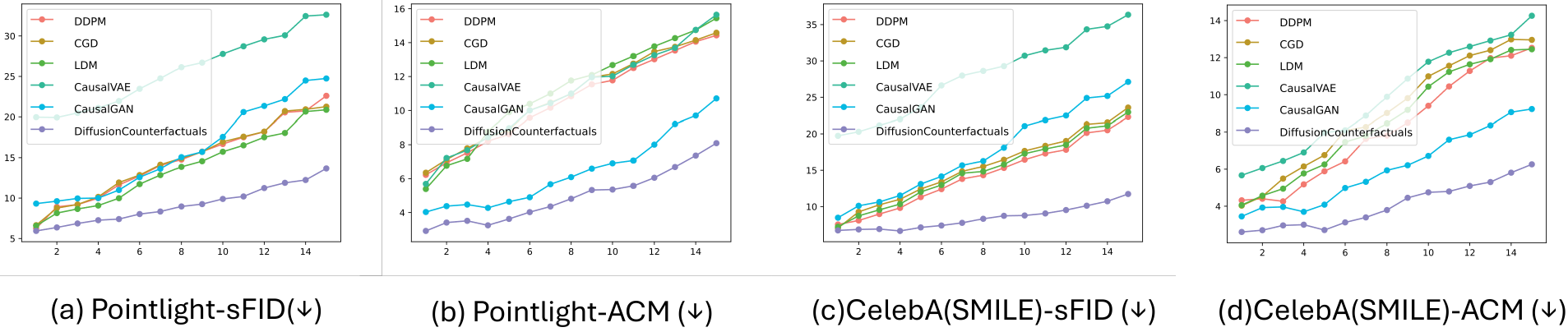}
    \caption{Sequential counterfactual generations results on Shadow-Pointlight and CelebA(SMILE). We evaluate the sequential counterfactual generated samples in an autoregressive way.}
    \label{fig:squential3}
    \vspace{-\baselineskip}
\end{figure*}

\begin{figure*}
    \centering
     \includegraphics[width=0.7\textwidth]{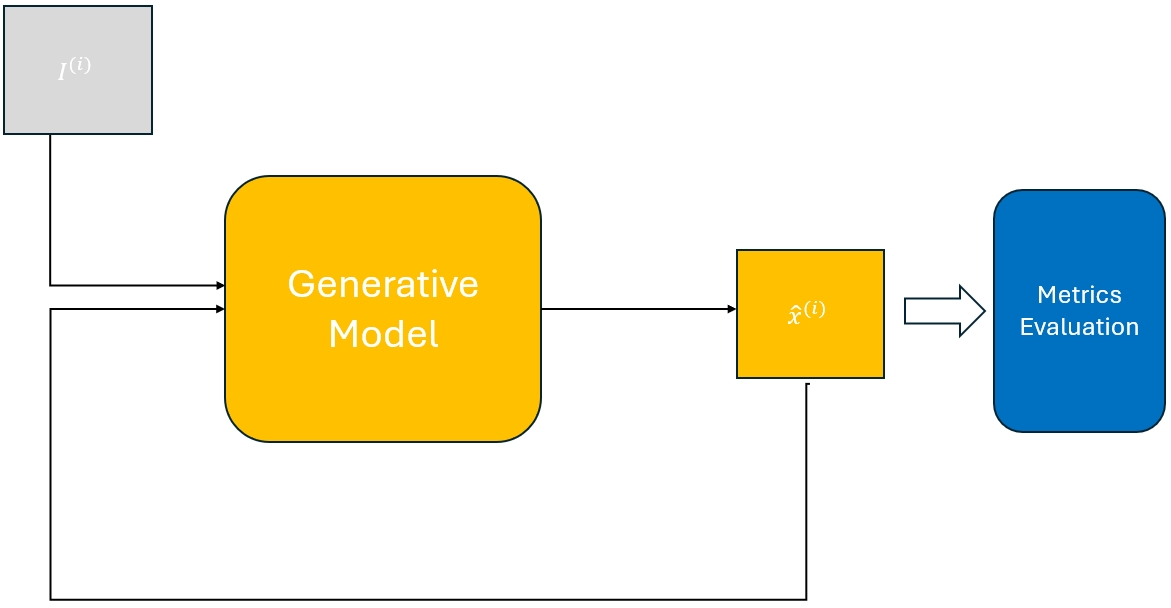}
    \caption{Sequential counterfactual samples generation evaluation pipeline, where the generated sample is fed into the generation model to generate next counterfactual sample in an autogressive manner.}
    \label{fig:autogressive}
    \vspace{-\baselineskip}
\end{figure*}

As described in \cref{sec:compare}, we include sequential generation results on Shadow-Sunlight and CelebA(BEARD). In this section, we include the autogressive evaluation discussed in \cref{sec:compare} as \cref{fig:autogressive}. Further, we include the results on other datasets in 
\cref{fig:squential2,fig:squential3}.


As shown in \cref{fig:squential,fig:squential2,fig:squential3}, our method consistently generates high-quality counterfactual samples across multiple steps in the sequence. The generated samples exhibit a high degree of consistency and coherence, accurately reflecting the expected changes based on the interventions applied to the causal factors. This highlights the effectiveness of our approach in capturing and leveraging causal knowledge in the generation process.
In contrast, baseline methods that do not incorporate causal knowledge struggle to maintain consistency among the generated samples in the sequence. The quality and coherence of the counterfactual samples tend to degrade as the number of steps increases, leading to inconsistencies and artifacts in the generated images. This underscores the importance of using causal knowledge in the generation process to obtain coherent and accurate results.
The ability to generate consistent and coherent counterfactual samples across multiple steps has significant implications for various applications, such as planning, decision-making, and scenario analysis. By incorporating causal knowledge, our method enables the generation of plausible and realistic counterfactual scenarios, providing valuable insights into the potential outcomes of different interventions and actions.
Overall, the comparison of sequential counterfactual sample generation demonstrates the superiority of our approach in terms of consistency, coherence, and accuracy. By leveraging causal knowledge, our method produces high-quality counterfactual samples that maintain their integrity across multiple steps, outperforming baseline methods that do not consider causal relationships.

\section{Implementation Details}
\label{app:implementation_details}
For each dataset, the diffusion network $g_\theta$ is constructed as an Unet-like architecture \cite{unet}, with skip-connections that enable efficient information flow between the encoder and decoder layers \cite{ddpm}. The causal projector $h_{\phi}$, responsible for mapping the learned representations to the causal factors, consists of the encoder from $g_\theta$, followed by a pooling layer and a linear regressor. For the neural structural causal model (NSCM), we use the same architecture with \cite{deci}.
All models incorporate time conditioning, where time is represented as a scalar value. To effectively embed this temporal information, we employ the sinusoidal position embedding technique used in transformers \citep{Vaswani2017AttentionNeed}. The embedded time is then integrated into the convolutional models through an Adaptive Group Normalization layer, which is applied within each residual block \cite{Nichol2021ImprovedModels}. The architectural design and training methodology of our models are inspired by the work of \cite{ddpm}. We optimize all models using the Adam optimizer with $\beta_1=0.9$ and $\beta_2=0.999$. To ensure computational efficiency, we train our models in 16-bit precision using loss-scaling techniques while maintaining 32-bit precision for weights, exponential moving average (EMA), and optimizer state. An EMA rate of 0.9999 is consistently used across all experiments.
For sampling, we employ the Denoising Diffusion Implicit Models (DDIM) \cite{ddim} approach with 100 timesteps throughout our experiments. For dataset with image size $96\times 96$, the experiments are run on one 1080Ti GPU, and for dataset with image size $128 \times 128$, the experiments are turn on two 1080Ti GPUs.

\begin{table}[ht]
    \setlength\tabcolsep{4pt}
    \begin{center}
    \caption{Hyperparameters for models.}
    \label{tab:hyperparameter}
    \begin{small}
    \begin{tabular}{l||cc|cc}
     image size &  \multicolumn{2}{c|}{$128\times 128$} & \multicolumn{2}{c}{$96\times 96$} \\
    \midrule
    model & $g_\theta$ & $h_{\phi}$ & $g_\theta$ & $h_\phi$ \\
    Diffusion steps & 1000 & 1000 & 1000 & 1000 \\
    Model size & 5M  & 2M  & 2M  & 1M\\
    Channels & 128 & 64 & 64 & 32 \\
    Depth &  2 &  2  &  1 &  1\\
    Channels multiple & 1,2,4 & 1,2,4 & 1,2,4 & 1,2,4,4 \\
    Attention resolution &  16,8 &  16,8 &  -  &  -  \\
    Batch size & 64 & 64 & 128 & 128 \\
    Iterations & $1.2$M & $800$K   & 800K & 50K  \\ 
    Learning Rate & 1e-4  & 1e-4 & 5e-4  & 5e-4\\
    \end{tabular}
    \end{small}
    \end{center}
     \vskip -0.2in
\end{table}

\section{Limitation and More Discussion}

While our proposed framework, DiffusionCounterfactuals, demonstrates substantial improvements in generating high-quality and causally consistent counterfactuals, several limitations need to be acknowledged: 

(1) The computational complexity and resources required for training diffusion models are substantial, particularly for high-dimensional data. This might limit the accessibility and practicality of our approach for researchers and practitioners with limited computational resources.

(2) Generalization to Diverse Domains:
While we have tested our framework on a variety of datasets, the generalizability of the proposed method to other domains or more complex real-world data remains to be fully validated. Further experiments on diverse and complex datasets are necessary to fully understand the robustness and applicability of our approach.

(3) Intervention Specificity:
   The accuracy of counterfactual generation is highly dependent on the specificity and correctness of the interventions applied. Inaccurate or poorly defined interventions may lead to less reliable counterfactuals.

We also listed the potential ethical concerns of this work.

(1) Bias and Fairness:
The method we propose might inadvertently learn and perpetuate biases present in the training data. This can lead to biased counterfactual inferences, especially in sensitive applications such as healthcare or criminal justice, where fairness is paramount.

(2) Data Privacy:
The use of high-dimensional and potentially sensitive data necessitates strict adherence to data privacy standards. Ensuring that our models and the training process comply with data protection regulations is crucial to prevent unauthorized access or misuse of data.
\clearpage 

\section{Visualization Results}
In this section, we first show the visualization results on datasets when intervening on specific the cause factors. Then, we include more results where each original sample are randomly intervened on cause factors.

\begin{figure*}
    \centering
\includegraphics[width=0.93\textwidth]{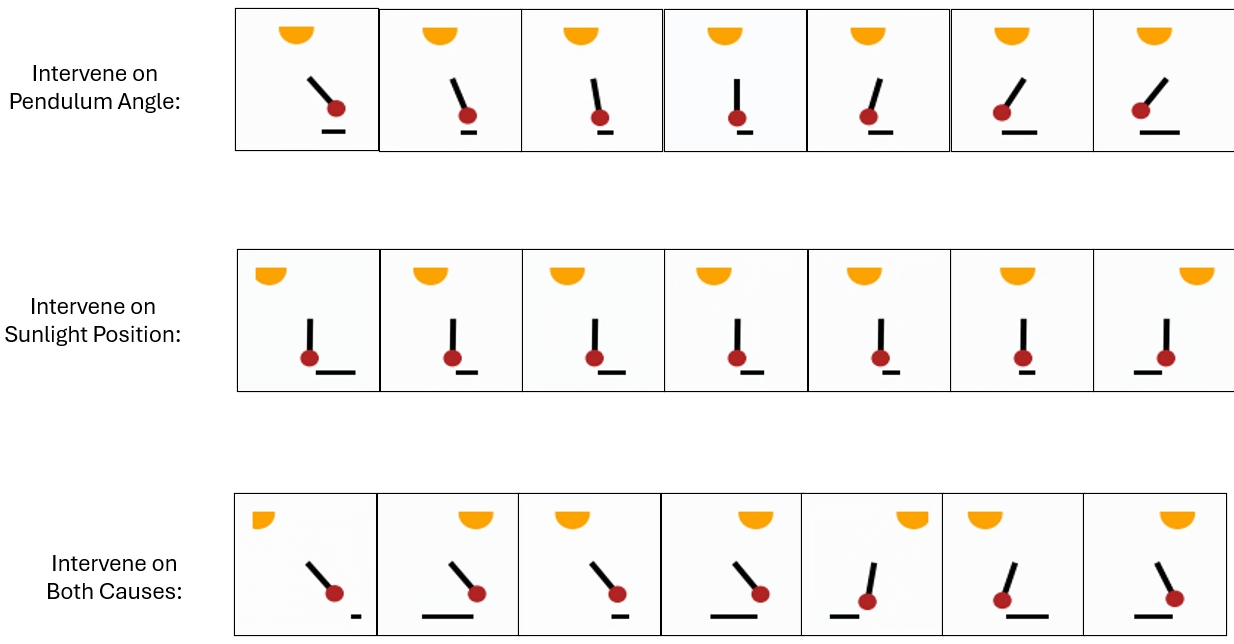}
    \caption{Pendulum counterfactual inference visualization when intervening on different cause factors.}
    \label{fig:pendulum-counterfactuals}
    \vspace{-\baselineskip}
\end{figure*}
\begin{figure*}
    \centering
\includegraphics[width=0.93\textwidth]{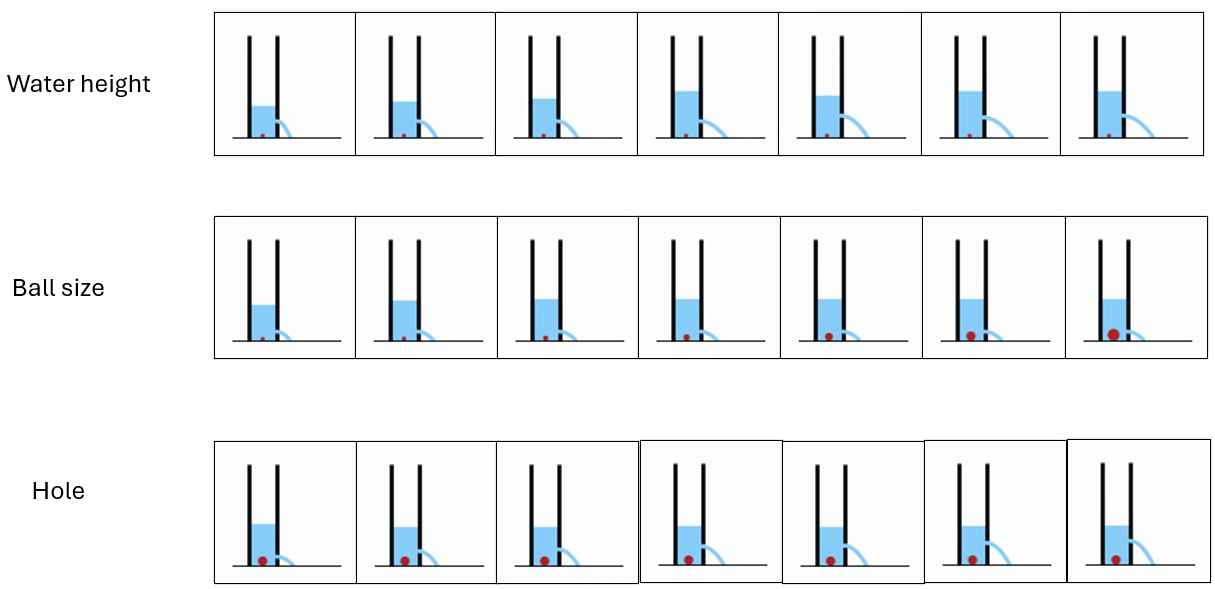}
    \caption{Flow counterfactual inference visualization when intervening on different cause factors.}
    \label{fig:flow-counterfactuals}
    \vspace{-\baselineskip}
\end{figure*}
\begin{figure*}
    \centering
\includegraphics[width=0.93\textwidth]{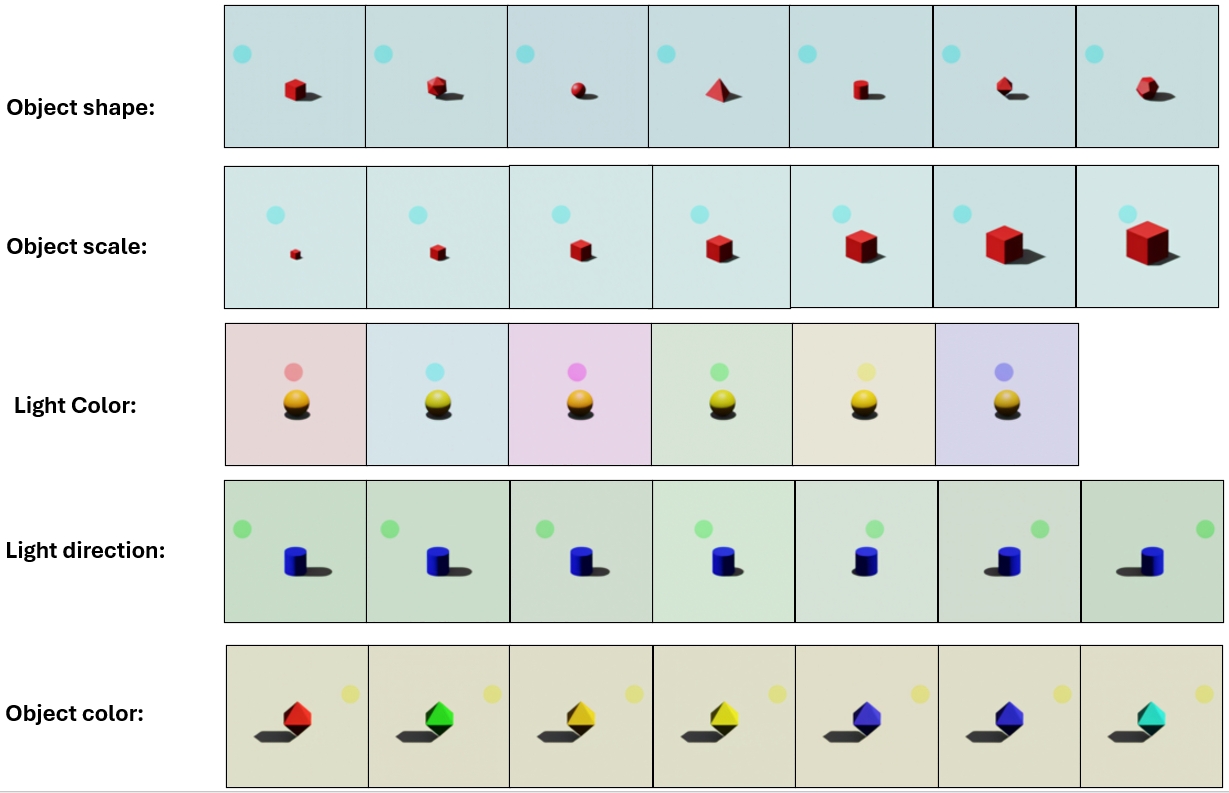}
    \caption{Shadow-Sunlight counterfactual inference visualization when intervening on different cause factors.}
    \label{fig:sunlight-counterfactuals}
    \vspace{-\baselineskip}
\end{figure*}
\begin{figure*}
    \centering
\includegraphics[width=0.93\textwidth]{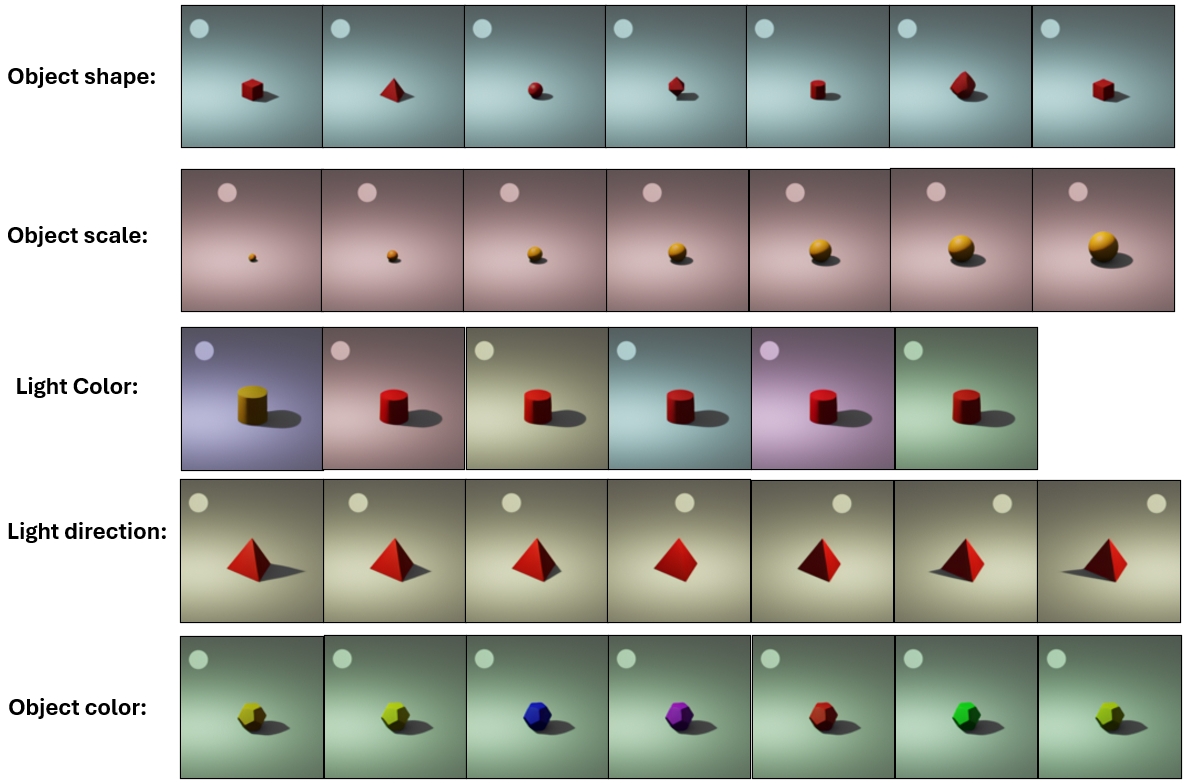}
    \caption{Shadow-Pointlight counterfactual inference visualization when intervening on different cause factors.}
    \label{fig:pointlight-counterfactuals}
    \vspace{-\baselineskip}
\end{figure*}

\begin{figure*}
    \centering
     \includegraphics[width=0.93\textwidth]{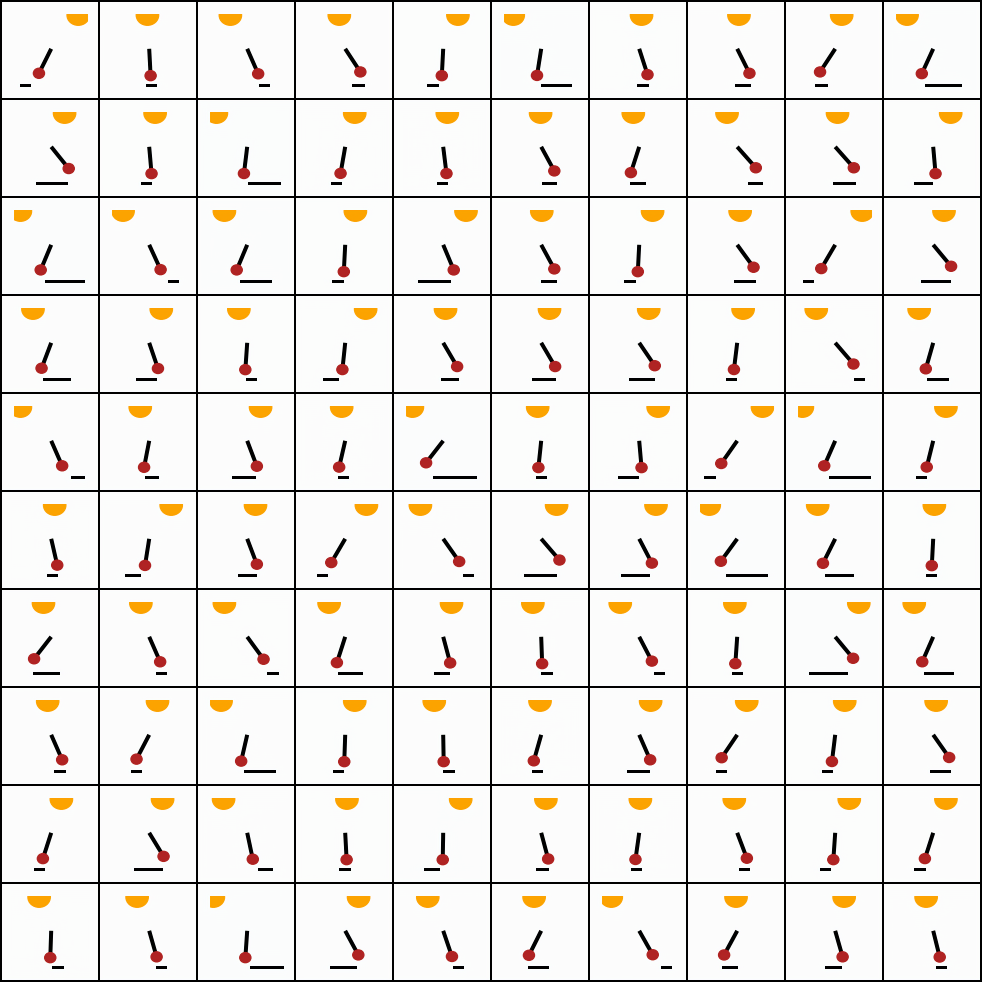}
    \caption{More Pendulum counterfactual inference results.}
    \label{fig:pendulum-all}
    \vspace{-\baselineskip}
\end{figure*}
\begin{figure*}
    \centering
     \includegraphics[width=0.93\textwidth]{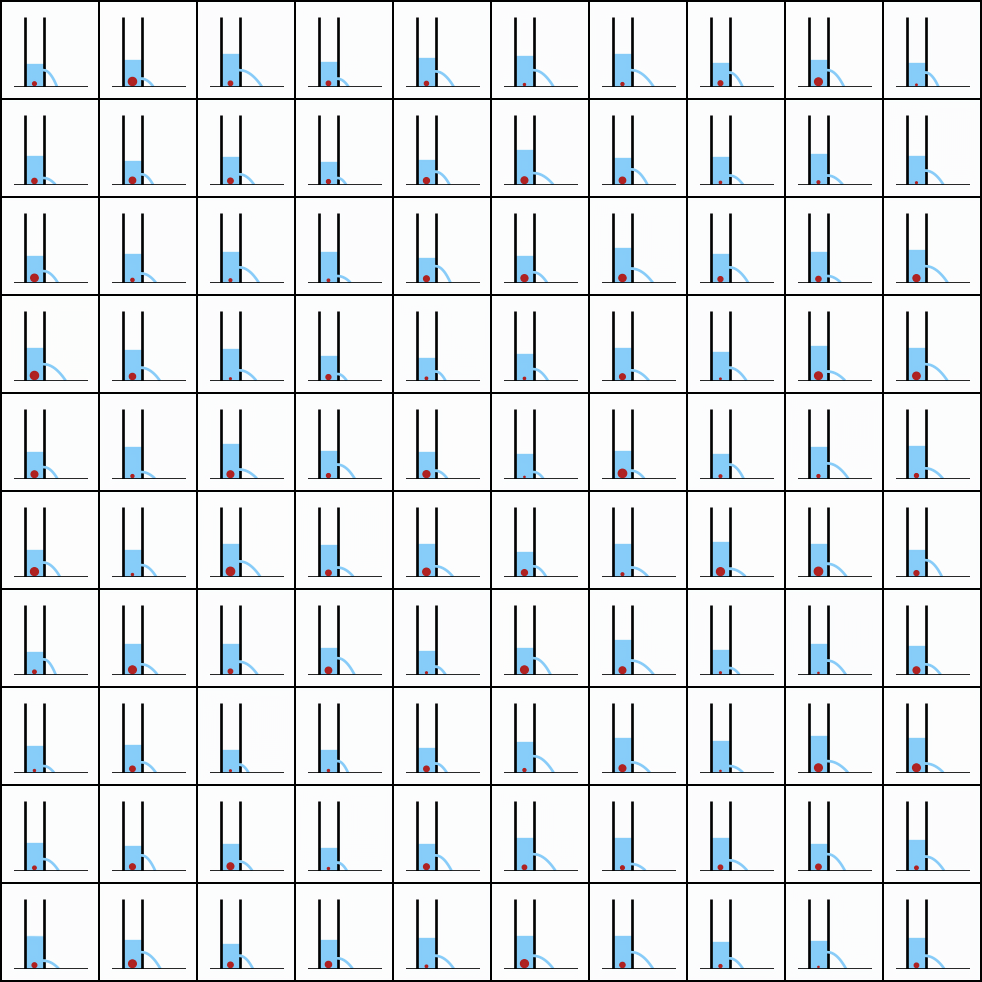}
    \caption{More Flow dataset counterfactual inference results.}
    \label{fig:flow-all}
    \vspace{-\baselineskip}
\end{figure*}
\begin{figure*}
    \centering
     \includegraphics[width=0.93\textwidth]{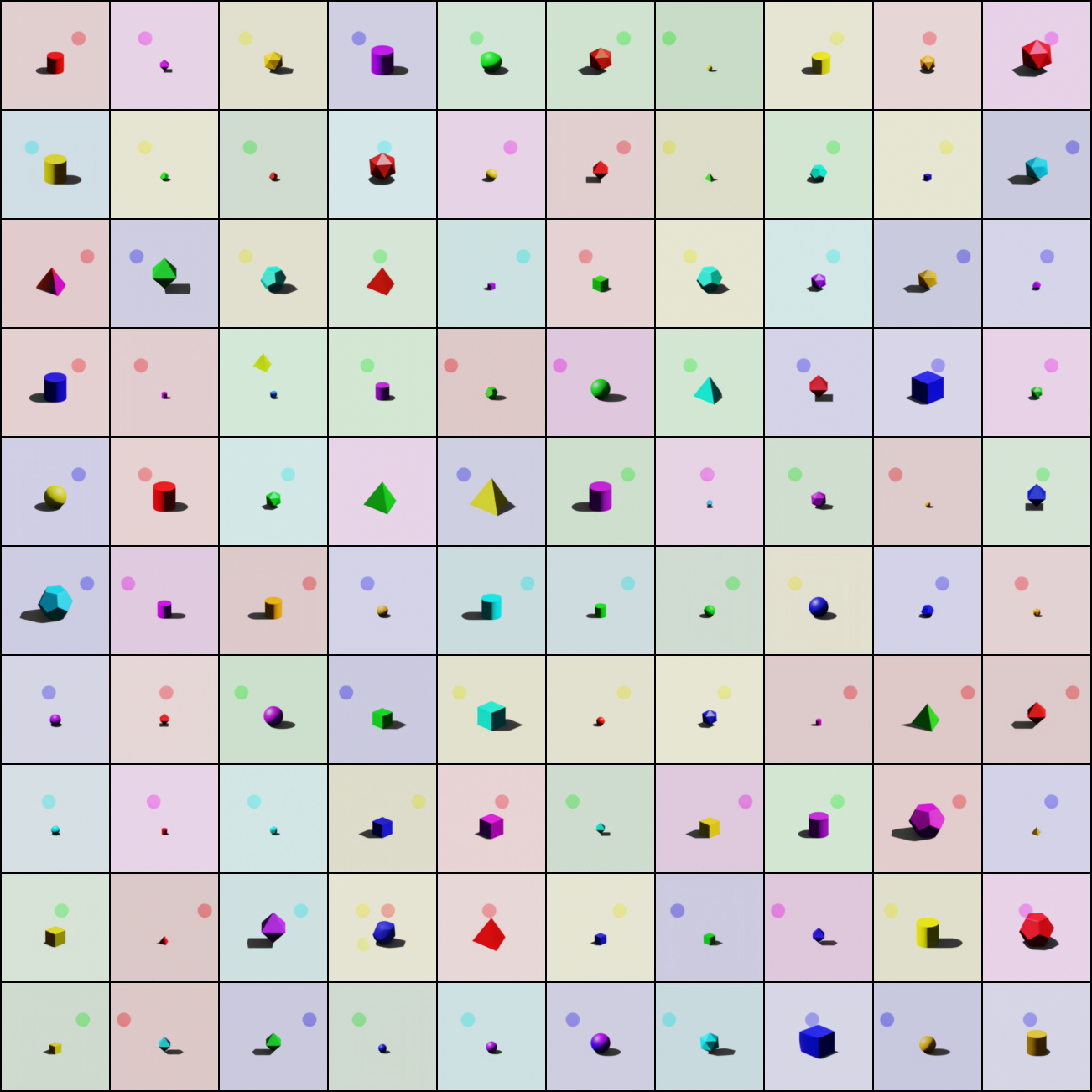}
    \caption{More Shadow-Sunlight counterfactual inference results.}
    \label{fig:sunlight-all}
    \vspace{-\baselineskip}
\end{figure*}
\begin{figure*}
    \centering
     \includegraphics[width=0.93\textwidth]{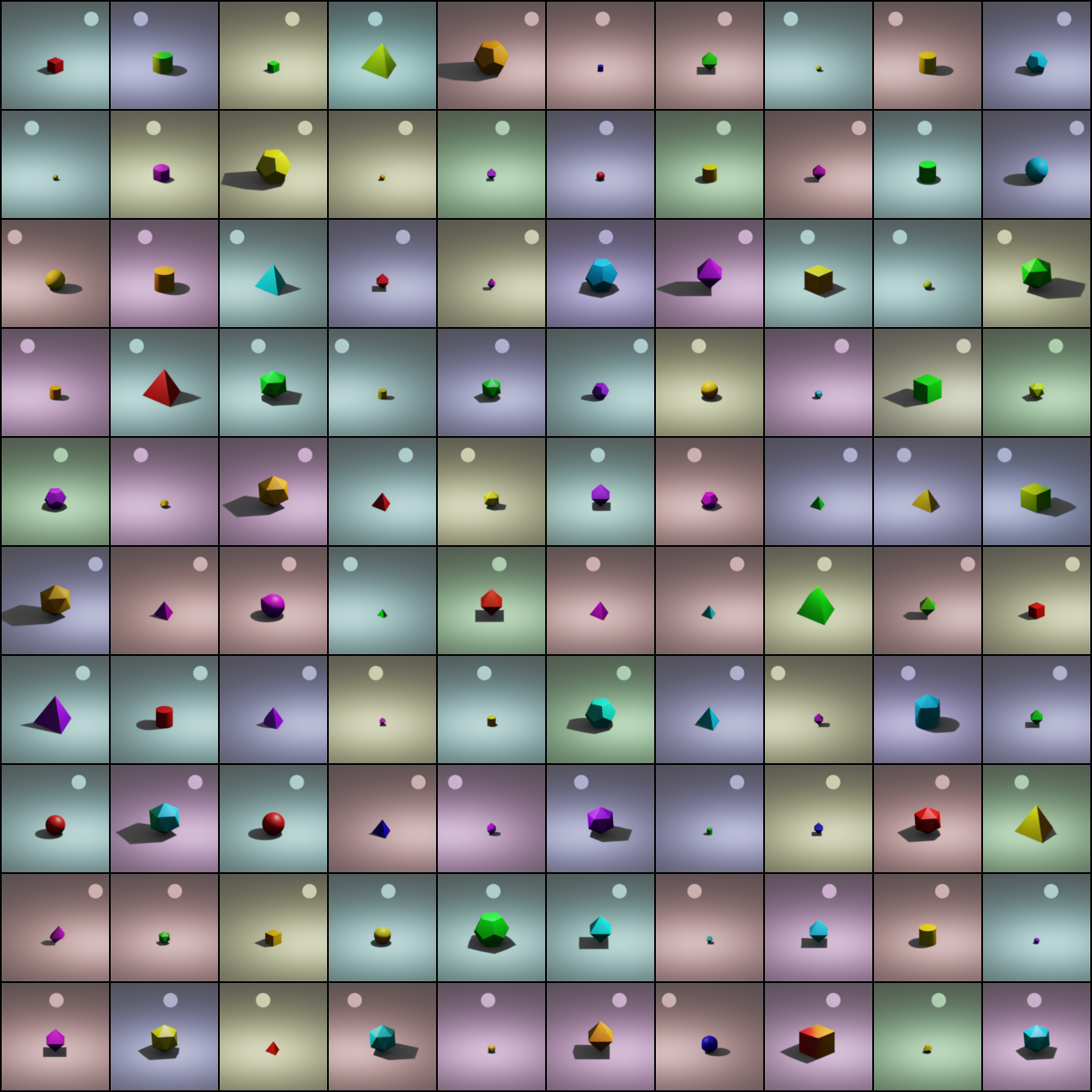}
    \caption{More Shadow-Pointlight counterfactual inference results.}
    \label{fig:pointlight-all}
    \vspace{-\baselineskip}
\end{figure*}
\begin{figure*}
    \centering
     \includegraphics[width=0.93\textwidth]{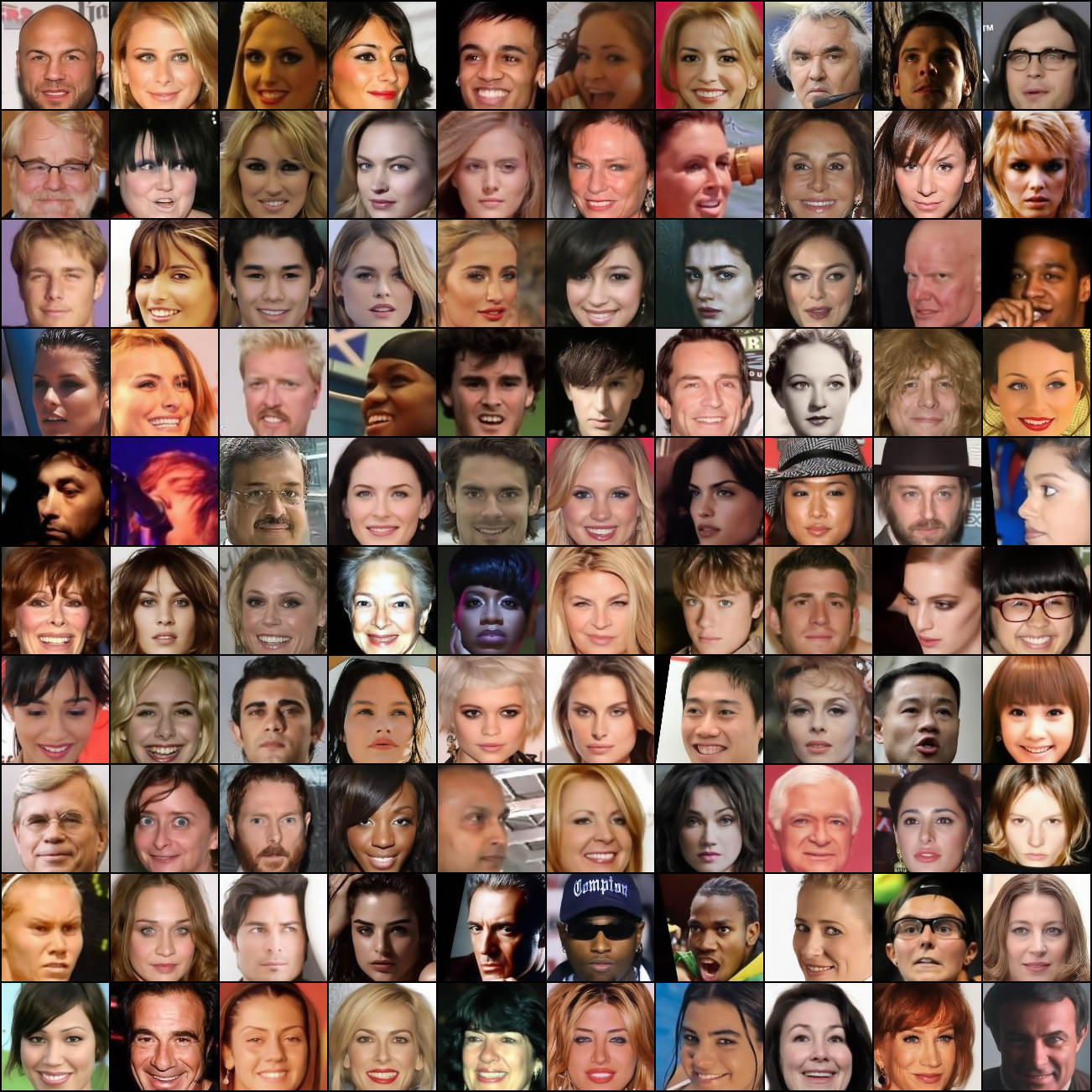}
    \caption{More CelebA(BEARD) counterfactual inference results.}
    \label{fig:beard-all}
    \vspace{-\baselineskip}
\end{figure*}
\begin{figure*}
    \centering
     \includegraphics[width=0.93\textwidth]{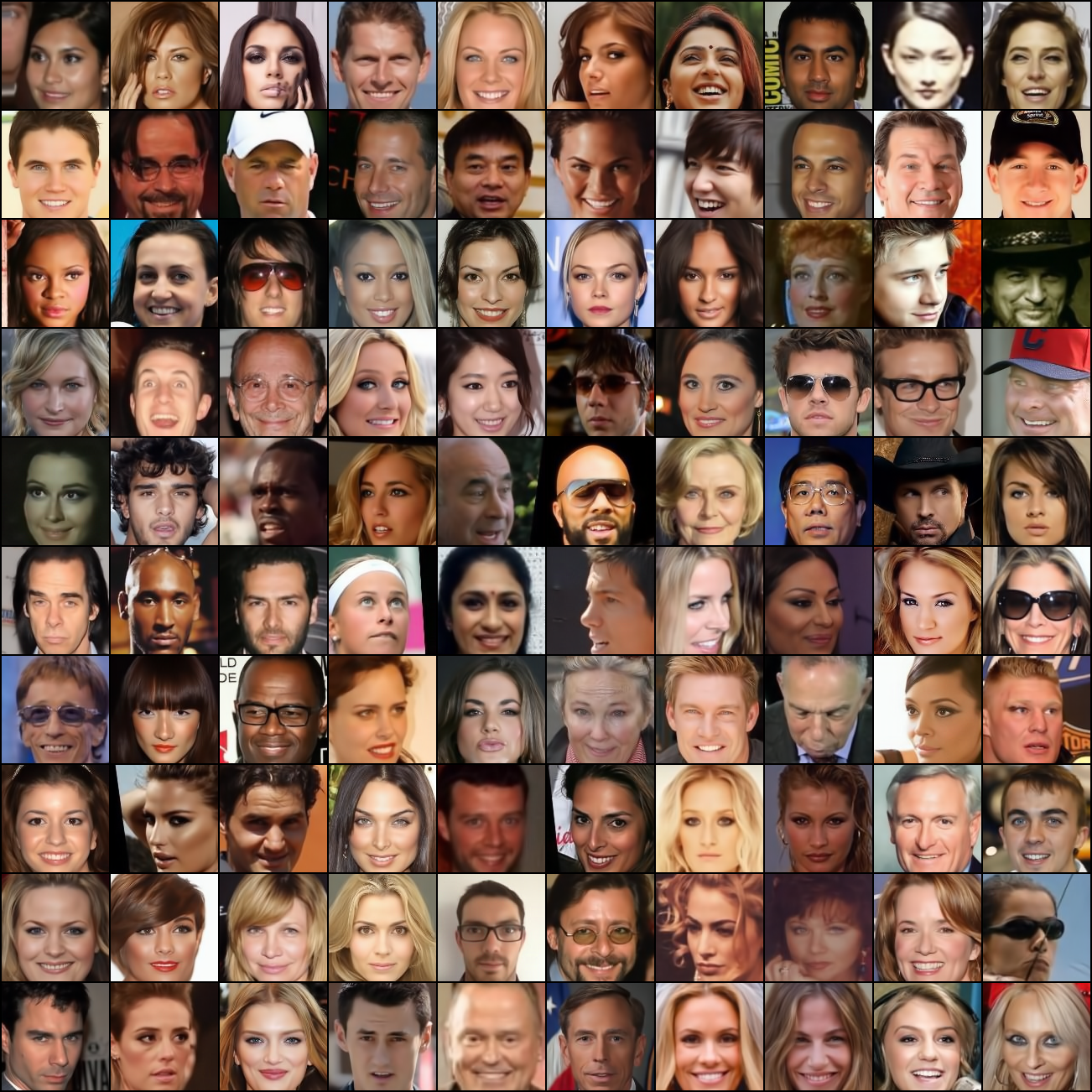}
    \caption{More CelebA(SMILE) counterfactual inference results.}
    \label{fig:smile-all}
    \vspace{-\baselineskip}
\end{figure*}

\end{document}